%% file: main.tex
\documentclass[10pt,twocolumn,letterpaper]{article}

\usepackage{wacv}

\makeatletter
\renewcommand\paragraph{\@startsection{paragraph}{4}{\z@}{.5ex}{-1em}{\normalfont\normalsize\bfseries}}
\makeatother
\input{tex/plots}

\usepackage{times}
\usepackage{epsfig}
\usepackage{graphicx}
\usepackage{amsmath}
\usepackage{amssymb}
\usepackage{booktabs}

\usepackage{comment}
\usepackage{color}
\usepackage[accsupp]{axessibility}  

\usepackage[utf8]{inputenc}
\usepackage{kotex}
\usepackage{kotex-logo}
\usepackage{mathtools}
\usepackage{enumitem}
\usepackage{xspace}
\usepackage{breakcites}
\usepackage{contour}
\usepackage{pifont}
\usepackage{mycommands}

\wacvfinalcopy 

\ifwacvfinal
\usepackage[breaklinks=true,bookmarks=false]{hyperref}
\else
\usepackage[pagebackref=true,breaklinks=true,colorlinks,bookmarks=false]{hyperref}
\fi

\begin{document}

\title{Boosting vision transformers for image retrieval}

\author{
Chull Hwan Song$^1$ \ \ \ \ Jooyoung Yoon$^1$ \ \ \ \ Shunghyun Choi$^1$ \ \ \ \ Yannis Avrithis$^{2,3}$\\
\normalsize $^1$Dealicious Inc.\quad
\normalsize $^2$Institute of Advanced Research on Artificial Intelligence (IARAI)\quad
\normalsize $^3$Athena RC\\
}

\maketitle
\thispagestyle{empty}

\newcommand{\Note}[3]{{\color{#1}[#2: #3]}}
\newcommand{\iavr}[1]{\Note{blue}{Y}{#1}}
\newcommand{\song}[1]{\Note{orange}{CH}{#1}}
\newcommand{\schoi}[1]{\Note{teal}{S}{#1}}
\definecolor{airforceblue}{rgb}{0.36, 0.54, 0.66}
\newcommand{\yoon}[1]{\Note{airforceblue}{JY}{#1}}

\input{tex/abbrev}
\input{tex/defn}

\begin{abstract}
   Vision transformers have achieved remarkable progress in vision tasks such as image classification and detection.
   However, in instance-level image retrieval, transformers have not yet shown good performance compared to convolutional networks. We propose a number of improvements that make transformers outperform the state of the art for the first time. (1) We show that a hybrid architecture is more effective than plain transformers, by a large margin. (2) We introduce two branches collecting global (classification token) and local (patch tokens) information, from which we form a global image representation. (3) In each branch, we collect multi-layer features from the transformer encoder, corresponding to skip connections across distant layers. (4) We enhance locality of interactions at the deeper layers of the encoder, which is the relative weakness of vision transformers. We train our model on all commonly used training sets and, for the first time, we make fair comparisons separately per training set. In all cases, we outperform previous models based on global representation. Public code is available at \url{https://github.com/dealicious-inc/DToP}.
\end{abstract}

\input{tex/intro}

\input{tex/method}

\input{tex/exp-setup}
\input{tex/exp-bench}
\input{tex/exp-vis}
\input{tex/exp-ablation}

\section{Conclusion}

We have introduced a new approach to leverage the power of vision transformers for image retrieval, reaching or setting new state of the art with global representations for the first time. By collecting both global and local features from multiple layers and enhancing locality in the last layers, we learn powerful representations that have improved localization properties, in the sense of attending the object of interest. Some of our ideas may be applicable beyond transformers and beyond retrieval. Our unexpected finding that a hybrid architecture outperforms most transformers by a large margin may open new directions in architecture design. A very interesting future direction is self-supervised learning of transformers for retrieval.

{\small
\bibliographystyle{ieee_fullname}
\bibliography{main}
}

\input{tex/supp}

\end{document}

%% file: tex/plots.tex
\makeatletter
\@namedef{ver@everyshi.sty}{}
\makeatother

\usepackage[dvipsnames,svgnames,x11names]{xcolor}
\usepackage{tikz}
\usetikzlibrary{arrows.meta,shapes,calc,matrix,fit,positioning,backgrounds,decorations.markings}
\usepackage{pgfplots}
\usepackage{pgfplotstable}
\pgfplotsset{compat=1.9}
\usepackage{xstring}

\usepgfplotslibrary{external}
\newcommand{\extfig}[2]{\tikzsetnextfilename{#1}{#2}}

\IfBeginWith*{\jobname}{fig/extern/}{\finalcopy}{}

\tikzstyle{every picture}+=[
	remember picture,
	every text node part/.style={align=center},
	every matrix/.append style={ampersand replacement=\&},
]
\tikzstyle{tight} = [inner sep=0pt,outer sep=0pt]
\tikzstyle{node}  = [draw,circle,tight,minimum size=12pt,anchor=center]
\tikzstyle{op}    = [draw,circle,tight]
\tikzstyle{dot}   = [fill,draw,circle,inner sep=1pt,outer sep=0]
\tikzstyle{pt}    = [fill,draw,circle,inner sep=1.5pt,outer sep=.2pt]
\tikzstyle{box}   = [draw,rectangle,inner sep=3pt]
\tikzstyle{high}  = [black!60]
\tikzstyle{group} = [high,box,opacity=.5]
\tikzstyle{dim1}  = [fill opacity=.3,text opacity=1]
\tikzstyle{dim2}  = [fill opacity=.5,text opacity=1]
\tikzstyle{dim3}  = [fill opacity=.7,text opacity=1]
\tikzstyle{rectc} = [tight,transform shape]
\tikzstyle{rect}  = [rectc,anchor=south west]

\newcommand{\leg}[1]{\addlegendentry{#1}}

\tikzset{every mark/.append style={solid}}
\pgfplotsset{
	grid=both, width=\columnwidth, try min ticks=5,
	every axis/.append style={font=\small},
	every axis plot/.append style={thick,mark=none,mark size=1.8,tension=0.18},
	legend cell align=left, legend style={fill opacity=0.8},
	xticklabel={\pgfmathprintnumber[assume math mode=true]{\tick}},
	yticklabel={\pgfmathprintnumber[assume math mode=true]{\tick}},
	nodes near coords math/.style={
		nodes near coords={\pgfmathprintnumber[assume math mode=true]{\pgfplotspointmeta}},
	},
}

\pgfplotsset{
	dash/.style={mark=o,dashed,opacity=0.6},
	dott/.style={mark=o,dotted,opacity=0.6},
	nolim/.style={enlargelimits=false},
	plain/.style={every axis plot/.append style={},nolim,grid=none},
}

\pgfdeclarelayer{bg4}
\pgfdeclarelayer{bg3}
\pgfdeclarelayer{bg2}
\pgfdeclarelayer{bg1}
\pgfdeclarelayer{fg1}
\pgfdeclarelayer{fg2}
\pgfdeclarelayer{fg3}
\pgfdeclarelayer{fg4}
\pgfsetlayers{bg4,bg3,bg2,bg1,main,fg1,fg2,fg3,fg4}

\pgfkeys{/tikz/.cd, aspect/.store in=\aspect, aspect=1}
\pgfkeys{/tikz/.cd, depth/.store in=\depth, depth=.5}
\pgfkeys{/tikz/.cd, stepx/.store in=\step, stepx=1}

\tikzstyle{geom} = [line join=bevel,aspect=1,depth=.5,z={(\depth*\aspect,\depth)}]
\tikzstyle{wire} = [geom,draw,thick]

\def\cx[#1,#2,#3]{#1}
\def\cy[#1,#2,#3]{#2}
\def\cz[#1,#2,#3]{#3}
\def\ex[#1,#2,#3]{#1,0,0}
\def\ey[#1,#2,#3]{0,#2,0}
\def\ez[#1,#2,#3]{0,0,#3}



%% file: tex/abbrev.tex
\newcommand{\head}[1]{{\smallskip\noindent\textbf{#1}}}
\newcommand{\alert}[1]{{\color{red}{#1}}}
\newcommand{\sm}{\scriptsize}
\newcommand{\eq}[1]{(\ref{eq:#1})}

\newcommand{\Th}[1]{\textsc{#1}}
\newcommand{\mr}[2]{\multirow{#1}{*}{#2}}
\newcommand{\mc}[2]{\multicolumn{#1}{c}{#2}}
\newcommand{\mca}[3]{\multicolumn{#1}{#2}{#3}}
\newcommand{\tb}[1]{\textbf{#1}}
\newcommand{\ch}{\checkmark}

\newcommand{\red}[1]{{\color{red}{#1}}}
\newcommand{\blue}[1]{{\color{blue}{#1}}}
\newcommand{\green}[1]{\color{green}{#1}}
\newcommand{\gray}[1]{{\color{gray}{#1}}}
\newcommand{\orange}[1]{{\color{orange}{#1}}}

\newcommand{\citeme}[1]{\red{[XX]}}
\newcommand{\refme}[1]{\red{(XX)}}

\newcommand{\fig}[2][1]{\includegraphics[width=#1\linewidth]{fig/#2}}
\newcommand{\figh}[2][1]{\includegraphics[height=#1\linewidth]{fig/#2}}


\newcommand{\tran}{^\top}
\newcommand{\mtran}{^{-\top}}
\newcommand{\zcol}{\mathbf{0}}
\newcommand{\zrow}{\zcol\tran}

\newcommand{\ind}{\mathbbm{1}}
\newcommand{\expect}{\mathbb{E}}
\newcommand{\nat}{\mathbb{N}}
\newcommand{\zahl}{\mathbb{Z}}
\newcommand{\real}{\mathbb{R}}
\newcommand{\proj}{\mathbb{P}}
\newcommand{\prob}{\mathbf{Pr}}
\newcommand{\normal}{\mathcal{N}}

\newcommand{\mif}{\textrm{if}\ }
\newcommand{\other}{\textrm{otherwise}}
\newcommand{\minimize}{\textrm{minimize}\ }
\newcommand{\maximize}{\textrm{maximize}\ }
\newcommand{\st}{\textrm{subject\ to}\ }

\newcommand{\id}{\operatorname{id}}
\newcommand{\const}{\operatorname{const}}
\newcommand{\sgn}{\operatorname{sgn}}
\newcommand{\var}{\operatorname{Var}}
\newcommand{\mean}{\operatorname{mean}}
\newcommand{\trace}{\operatorname{tr}}
\newcommand{\diag}{\operatorname{diag}}
\newcommand{\vect}{\operatorname{vec}}
\newcommand{\cov}{\operatorname{cov}}
\newcommand{\sign}{\operatorname{sign}}
\newcommand{\prj}{\operatorname{proj}}

\newcommand{\sigmoid}{\operatorname{sigmoid}}
\newcommand{\softmax}{\operatorname{softmax}}
\newcommand{\clip}{\operatorname{clip}}

\newcommand{\defn}{\mathrel{:=}}
\newcommand{\peq}{\mathrel{+\!=}}
\newcommand{\meq}{\mathrel{-\!=}}

\newcommand{\floor}[1]{\left\lfloor{#1}\right\rfloor}
\newcommand{\ceil}[1]{\left\lceil{#1}\right\rceil}
\newcommand{\inner}[1]{\left\langle{#1}\right\rangle}
\newcommand{\norm}[1]{\left\|{#1}\right\|}
\newcommand{\abs}[1]{\left|{#1}\right|}
\newcommand{\frob}[1]{\norm{#1}_F}
\newcommand{\card}[1]{\left|{#1}\right|\xspace}
\newcommand{\diff}{\mathrm{d}}
\newcommand{\der}[3][]{\frac{d^{#1}#2}{d#3^{#1}}}
\newcommand{\pder}[3][]{\frac{\partial^{#1}{#2}}{\partial{#3^{#1}}}}
\newcommand{\ipder}[3][]{\partial^{#1}{#2}/\partial{#3^{#1}}}
\newcommand{\dder}[3]{\frac{\partial^2{#1}}{\partial{#2}\partial{#3}}}

\newcommand{\wb}[1]{\overline{#1}}
\newcommand{\wt}[1]{\widetilde{#1}}

\def\xssp{\hspace{1pt}}
\def\ssp{\hspace{3pt}}
\def\msp{\hspace{5pt}}
\def\lsp{\hspace{12pt}}

\newcommand{\cA}{\mathcal{A}}
\newcommand{\cB}{\mathcal{B}}
\newcommand{\cC}{\mathcal{C}}
\newcommand{\cD}{\mathcal{D}}
\newcommand{\cE}{\mathcal{E}}
\newcommand{\cF}{\mathcal{F}}
\newcommand{\cG}{\mathcal{G}}
\newcommand{\cH}{\mathcal{H}}
\newcommand{\cI}{\mathcal{I}}
\newcommand{\cJ}{\mathcal{J}}
\newcommand{\cK}{\mathcal{K}}
\newcommand{\cL}{\mathcal{L}}
\newcommand{\cM}{\mathcal{M}}
\newcommand{\cN}{\mathcal{N}}
\newcommand{\cO}{\mathcal{O}}
\newcommand{\cP}{\mathcal{P}}
\newcommand{\cQ}{\mathcal{Q}}
\newcommand{\cR}{\mathcal{R}}
\newcommand{\cS}{\mathcal{S}}
\newcommand{\cT}{\mathcal{T}}
\newcommand{\cU}{\mathcal{U}}
\newcommand{\cV}{\mathcal{V}}
\newcommand{\cW}{\mathcal{W}}
\newcommand{\cX}{\mathcal{X}}
\newcommand{\cY}{\mathcal{Y}}
\newcommand{\cZ}{\mathcal{Z}}

\newcommand{\vA}{\mathbf{A}}
\newcommand{\vB}{\mathbf{B}}
\newcommand{\vC}{\mathbf{C}}
\newcommand{\vD}{\mathbf{D}}
\newcommand{\vE}{\mathbf{E}}
\newcommand{\vF}{\mathbf{F}}
\newcommand{\vG}{\mathbf{G}}
\newcommand{\vH}{\mathbf{H}}
\newcommand{\vI}{\mathbf{I}}
\newcommand{\vJ}{\mathbf{J}}
\newcommand{\vK}{\mathbf{K}}
\newcommand{\vL}{\mathbf{L}}
\newcommand{\vM}{\mathbf{M}}
\newcommand{\vN}{\mathbf{N}}
\newcommand{\vO}{\mathbf{O}}
\newcommand{\vP}{\mathbf{P}}
\newcommand{\vQ}{\mathbf{Q}}
\newcommand{\vR}{\mathbf{R}}
\newcommand{\vS}{\mathbf{S}}
\newcommand{\vT}{\mathbf{T}}
\newcommand{\vU}{\mathbf{U}}
\newcommand{\vV}{\mathbf{V}}
\newcommand{\vW}{\mathbf{W}}
\newcommand{\vX}{\mathbf{X}}
\newcommand{\vY}{\mathbf{Y}}
\newcommand{\vZ}{\mathbf{Z}}

\newcommand{\va}{\mathbf{a}}
\newcommand{\vb}{\mathbf{b}}
\newcommand{\vc}{\mathbf{c}}
\newcommand{\vd}{\mathbf{d}}
\newcommand{\ve}{\mathbf{e}}
\newcommand{\vf}{\mathbf{f}}
\newcommand{\vg}{\mathbf{g}}
\newcommand{\vh}{\mathbf{h}}
\newcommand{\vi}{\mathbf{i}}
\newcommand{\vj}{\mathbf{j}}
\newcommand{\vk}{\mathbf{k}}
\newcommand{\vl}{\mathbf{l}}
\newcommand{\vm}{\mathbf{m}}
\newcommand{\vn}{\mathbf{n}}
\newcommand{\vo}{\mathbf{o}}
\newcommand{\vp}{\mathbf{p}}
\newcommand{\vq}{\mathbf{q}}
\newcommand{\vr}{\mathbf{r}}
\newcommand{\Vs}{\mathbf{s}}
\newcommand{\vt}{\mathbf{t}}
\newcommand{\vu}{\mathbf{u}}
\newcommand{\vv}{\mathbf{v}}
\newcommand{\vw}{\mathbf{w}}
\newcommand{\vx}{\mathbf{x}}
\newcommand{\vy}{\mathbf{y}}
\newcommand{\vz}{\mathbf{z}}

\newcommand{\vone}{\mathbf{1}}
\newcommand{\vzero}{\mathbf{0}}

\newcommand{\valpha}{{\boldsymbol{\alpha}}}
\newcommand{\vbeta}{{\boldsymbol{\beta}}}
\newcommand{\vgamma}{{\boldsymbol{\gamma}}}
\newcommand{\vdelta}{{\boldsymbol{\delta}}}
\newcommand{\vepsilon}{{\boldsymbol{\epsilon}}}
\newcommand{\vzeta}{{\boldsymbol{\zeta}}}
\newcommand{\veta}{{\boldsymbol{\eta}}}
\newcommand{\vtheta}{{\boldsymbol{\theta}}}
\newcommand{\viota}{{\boldsymbol{\iota}}}
\newcommand{\vkappa}{{\boldsymbol{\kappa}}}
\newcommand{\vlambda}{{\boldsymbol{\lambda}}}
\newcommand{\vmu}{{\boldsymbol{\mu}}}
\newcommand{\vnu}{{\boldsymbol{\nu}}}
\newcommand{\vxi}{{\boldsymbol{\xi}}}
\newcommand{\vomikron}{{\boldsymbol{\omikron}}}
\newcommand{\vpi}{{\boldsymbol{\pi}}}
\newcommand{\vrho}{{\boldsymbol{\rho}}}
\newcommand{\vsigma}{{\boldsymbol{\sigma}}}
\newcommand{\vtau}{{\boldsymbol{\tau}}}
\newcommand{\vupsilon}{{\boldsymbol{\upsilon}}}
\newcommand{\vphi}{{\boldsymbol{\phi}}}
\newcommand{\vchi}{{\boldsymbol{\chi}}}
\newcommand{\vpsi}{{\boldsymbol{\psi}}}
\newcommand{\vomega}{{\boldsymbol{\omega}}}

\newcommand{\rLambda}{\mathrm{\Lambda}}
\newcommand{\rSigma}{\mathrm{\Sigma}}

\newcommand{\vLambda}{\bm{\rLambda}}
\newcommand{\vSigma}{\bm{\rSigma}}

\makeatletter
\newcommand*\bdot{\mathpalette\bdot@{.7}}
\newcommand*\bdot@[2]{\mathbin{\vcenter{\hbox{\scalebox{#2}{$\m@th#1\bullet$}}}}}
\makeatother

\makeatletter
\DeclareRobustCommand\onedot{\futurelet\@let@token\@onedot}
\def\@onedot{\ifx\@let@token.\else.\null\fi\xspace}

\def\eg{\emph{e.g}\onedot} \def\Eg{\emph{E.g}\onedot}
\def\ie{\emph{i.e}\onedot} \def\Ie{\emph{I.e}\onedot}
\def\cf{\emph{cf}\onedot} \def\Cf{\emph{Cf}\onedot}
\def\etc{\emph{etc}\onedot} \def\vs{\emph{vs}\onedot}
\def\wrt{w.r.t\onedot} \def\dof{d.o.f\onedot} \def\aka{a.k.a\onedot}
\def\etal{\emph{et al}\onedot}
\makeatother

%% file: tex/defn.tex

\newcommand{\ours}{DToP\xspace}
\newcommand{\Ours}{\emph{deep token pooling} (\ours)\xspace}


\newcommand{\cls}{{\texttt{[CLS]}}\xspace}

\newcommand{\relu}{\operatorname{relu}}
\newcommand{\conv}{\operatorname{conv}}
\newcommand{\aconv}{\operatorname{aconv}}

\newcommand{\fc}{\textsc{fc}}
\newcommand{\gap}{\textsc{gap}}
\newcommand{\bn}{\textsc{bn}}
\newcommand{\dropout}{\textsc{dropout}}

\newcommand{\elm}{\textsc{elm}}
\newcommand{\irb}{\textsc{irb}}
\newcommand{\wav}{\textsc{wb}}
\newcommand{\aspp}{\textsc{aspp}}
\newcommand{\fuse}{\textsc{fuse}}


\def\oxf5k{Ox5k\xspace}
\def\paris6k{Par6k\xspace}
\def\roxf{$\cR$Oxford\xspace}
\def\rox{$\cR$Oxf\xspace}
\def\ro{$\cR$O\xspace}
\def\rpar{$\cR$Paris\xspace}
\def\rpa{$\cR$Par\xspace}
\def\rp{$\cR$P\xspace}
\def\r1m{$\cR$1M\xspace}
\def\rs{$\cR$100k\xspace}


\newcommand{\gain}[1]{{\color{green!60!black}#1}}

%% file: tex/intro.tex
\section{Introduction}
\label{sec:Introduction}

Instance-level image retrieval has undergone impressive progress in the deep learning era. Based on \emph{convolutional networks} (CNN), it is possible to learn compact, discriminative representations in either supervised or unsupervised settings. Advances concern mainly \emph{pooling methods}~\cite{Weyand02, Babenko03, Kalantidis01, Radenovi01, Gordo01}, loss functions originating in \emph{deep metric learning}~\cite{Gordo01,Radenovic01,Ng01}, large-scale open \emph{datasets}~\cite{Babenko01, Gordo01, Radenovic01, delf, Weyand01}, and \emph{competitions} such as Google landmark retrieval\footnote{https://www.kaggle.com/c/landmark-retrieval-2021}.

Studies of self-attention-based \emph{transformers}~\cite{att}, originating in the NLP field, have followed an explosive growth in computer vision too, starting with \emph{vision transformer} (ViT)~\cite{VIT}. However, most of these studies focus on image classification and detection. The few studies that are concerned with image retrieval~\cite{IRT, dino} find that transformers still underperform convolutional networks, even when trained on more data under better settings.

In this work, we study a large number of vision transformers on image retrieval and contribute ideas to improve their performance, without introducing a new architecture.
We are motivated by the fact that vision transformers may have a powerful built-in attention-based pooling mechanism, but this is learned on the training set distribution, while in image retrieval the test distribution is different. Hence, we need to go back to the patch token embeddings. We build a powerful global image representation by an advanced pooling mechanism over token embeddings from several of the last layers of the transformer encoder. We thus call our method \Ours.

Image retrieval studies are distinguished between \emph{global} and \emph{local} representations, involving one~\cite{Radenovic01, Ng01, yang2021dolg} and several~\cite{delf, cao2020unifying, Giorgos} vectors per image, respectively. We focus on the former as it is compact and allows simple and fast search. For the same reason, we do not focus on \emph{re-ranking}, based either on local feature geometry~\cite{delf, Oriane} or graph-based methods like diffusion~\cite{Donoser01, Iscen_2017_CVPR}.

We make the following contributions:
\begin{enumerate}[itemsep=2pt, parsep=0pt, topsep=0pt]
	\item We show the importance of inductive bias in the first layers for image retrieval.
	\item We handle dynamic image size at training.
	\item We collect global and local features from the classification and patch tokens respectively of multiple layers.
	\item We enhance locality of interactions in the last layers by means of lightweight, multi-scale convolution.
	\item We contribute to fair benchmarking by grouping results by training set and training models on all commonly used training sets in the literature.
	\item We achieve state of the art performance on image retrieval using transformers for the first time.
\end{enumerate}

%% file: tex/method.tex
\section{Method}
\label{sec:method}

\begin{figure*}[t]
\begin{center}
	\fig[1.0]{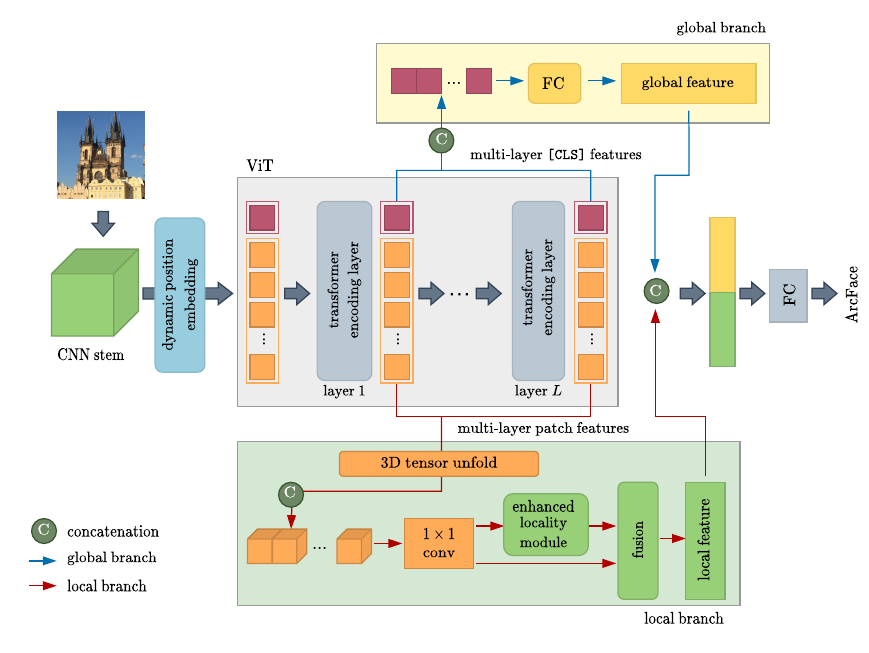}
\end{center}
\vspace{-30pt}
\caption{The high-level design of our \Ours. Using a transformer encoder (center), we build a global image representation for image retrieval by means of a \emph{global branch} (blue arrows, top) and a \emph{local branch} (red arrows, bottom), collecting \cls and patch token embeddings, respectively, from multiple layers. There are two mechanisms to improve locality of interactions (green): a CNN stem for the first layers (left), which amounts to a hybrid architecture, and our \emph{enhanced locality module} (ELM) (\autoref{fig:elm-dpe}(b)) in the local branch. Our \emph{dynamic position embedding} (DPE) (\autoref{fig:elm-dpe}(a)) allows for dynamic image size at training.}
\label{fig:arch}
\end{figure*}

\autoref{fig:arch} shows the proposed design of our \Ours.
We motivate and lay out its design principles in \autoref{sec:design}, discussing different components each time, after introducing the vision transformer in \autoref{sec:prelim}. We then provide a detailed account of the model in \autoref{sec:model}.

\subsection{Preliminaries: vision transformer}
\label{sec:prelim}

A transformer encoder, shown in the center of \autoref{fig:arch}, processes a sequence of token embeddings by allowing pairwise interactions in each layer. While we investigate a number of vision transformers, we follow ViT~\cite{VIT} here, which is our default choice. The \emph{input sequence} can be written as
\begin{equation}
	X = [\vx_\cls; \vx_1; \dots; \vx_M] \in \real^{(M+1) \times D},
\label{eq:input}
\end{equation}
where \emph{patch token} embeddings $\vx_1, \dots, \vx_M \in \real^D$ are obtained from the input image, the learnable \cls token embedding $\vx_\cls$ serves as global image representation at the output layer, $M$ is the sequence length and $D$ is the token embedding dimension.

There are two ways to form patch token embeddings. The most common is to decompose the input image into $M = wh$ raw, fixed-size, square non-overlapping patches and project them to $D$ dimensions via a learnable linear layer. Alternatively, one may use a convolutional network \emph{stem} to map the raw input image to a $w \times h \times D$ feature tensor, then fold this tensor into a sequence of $M = wh$ vectors of dimension $D$. This is called a \emph{hybrid} architecture. Here, $w \times h$ is \emph{input resolution}, \ie, the image resolution divided by the patch size in the first case or the downsampling ratio of the stem in the second.

The input sequence is added to a sequence of learnable \emph{position embeddings}, meant to preserve positional information, and given to the transformer encoder, which has $L$ layers preserving the sequence length and dimension. Each layer consists of a \emph{multi-head self attention} (MSA) and an MLP block. The output is the embedding of the [{\texttt{CLS}}] token at the last layer, $\vc^L$.


\subsection{Motivation and design principles}
\label{sec:design}

We are investigating a number of ideas, discussing related work in other tasks and laying out design principles accordingly. The overall goal is to use the features obtained by a vision transformer, without designing an entirely new architecture or extending an existing one too much.

\paragraph{Hybrid architecture}

As shown in the original ViT study~\cite{VIT}, hybrid models slightly outperform ViT at small computational budgets, but the difference vanishes for larger models. Of course, this finding refers to image classification tasks only. Although hybrid models are still studied~\cite{levit}, they are not mainstream: It is more common to introduce structure and inductive bias to transformer models themselves, where the input is still raw patches~\cite{Swin, pit, cvt, tokens, tnt}.

We are the first to conduct a large-scale investigation of different transformer architectures including hybrid models for image retrieval. Interestingly, we find that, in terms of global representation like the [{\texttt{CLS}}] token embeddings, the hybrid model originally introduced by~\cite{VIT} and consisting of a CNN stem and a ViT encoder performs best on image retrieval benchmarks by a large margin. As shown on the left in \autoref{fig:arch}, we use a CNN stem and a ViT encoder by default: The intermediate feature maps of the CNN stem are fed into ViT as token embeddings with patch size $1 \times 1$ rather than raw image patches.


\paragraph{Handling different image resolutions}

Image resolutions is an important factor in training image retrieval models. It is known that preserving original image resolution is effective~\cite{Hao2016WhatIT, Gordo01}. However, this leads to increased computational cost and longer training time. Focusing on image classification, MobileViT~\cite{mehta2021mobilevit} proposes a multi-scale sampler that randomly samples a spatial resolution from a fixed set and computes the batch size for this resolution at every training iteration. On image retrieval, \emph{group-size sampling}~\cite{Yokoo2020TwostageDR} has been shown very effective. Here, one constructs a mini batch with images of similar aspect ratios, resizing them to a prefixed size according to aspect ratio.

We follow this latter approach. However, because of different aspect ratio, the image size is still different per mini-batch, which presents a new challenge: Position embeddings are of fixed length, corresponding to fixed spatial resolution when unfolded. For this reason, as shown on the left in \autoref{fig:arch}, we propose \emph{dynamic position embedding} (DPE), whereby the fixed-size learned embeddings are dynamically resampled to the size of each mini-batch.


\paragraph{Global and local branches}

It is well known~\cite{delf,Radenovic01,Ng01} that an image retrieval model should focus on the target object, not the background. It is then no surprise that recent methods, focusing either on global or local representations, have a \emph{global} and a \emph{local branch} in their architecture after the backbone~\cite{cao2020unifying, Giorgos, wu2021learning, yang2021dolg, SCH01}. The objective of the local branch is to improve the localization properties of the model, even if the representation is eventually pooled into a single vector. Even though transformers have shown better localization properties than convolutional networks, especially in the self-supervised setting~\cite{dino,he2021masked,li2021mst}, the few studies so far on vision transformers for image retrieval are limited to using the [{\texttt{CLS}}] token from the last layer of ViT as a global representation~\cite{IRT, dino, gkelios2021investigating}.

In this context, our goal is to investigate the role of a local branch on top of a vision transformer encoder for image retrieval. This study is unique in that the local branch has access to patch token embeddings of different layers, re-introduces inductive bias by means of convolution at different scales and ends in global spatial pooling, thereby being complementary to the [{\texttt{CLS}}] token. As shown on the top/bottom in \autoref{fig:arch}, the global/local branch is based on the [{\texttt{CLS}}]/patch tokens, respectively. The final image representation is based on the concatenation of the features generated by the two branches.


\paragraph{Multi-layer features}

It is common in object detection, semantic segmentation and other dense prediction tasks to use features of different scales from different network layers, giving rise to \emph{feature pyramids}~\cite{ronneberger2015u, liu2016ssd, lin2017feature, li2018detnet, bifpn}. It is also common to introduce \emph{skip connections} within the architecture, sparsely or densely across layers, including architecture learning~\cite{huang2017densely, zhu2018sparsely,fang2020densely}. Apart from standard residual connections, connections across distant layers are not commonly studied in either image retrieval or vision transformers.

As shown on the top/bottom in \autoref{fig:arch}, without changing the encoder architecture itself, we investigate direct connections from several of its last layers to both the global and local branches, in the form of concatenation followed by a number of layers. This is similar to \emph{hypercolumns}~\cite{HAGM15}, but we are focusing on the last layers and building a global representation. The spatial resolution remains fixed in ViT, but we do take scale into account by means of dilated convolution. Interestingly, skip connections and especially direct connections to the output are known to improve the loss landscape of the network~\cite{li2018visualizing, nguyen2018loss}.


\paragraph{Enhancing locality}

The transformers mainly rely on global self-attention, which makes them good at modeling long-range dependencies. However, contrary to convolutional networks with fixed kernel size, they lack a mechanism to localize interactions. As a consequence, many studies~\cite{zhou2021ELSA, yuan2021incorporating, tnt, localvit, twins, conformer, chen2021regionvit} are proposed to improve ViT by bringing in locality.

In this direction, apart from using a CNN stem in the first layers, we introduce an \emph{enhanced locality module} (ELM) in the local branch, as shown in \autoref{fig:arch}. Our goal is to investigate inductive bias in the deeper layers of the encoder, without overly extending the architecture itself. For this reason, the design of ELM is extremely lightweight, inspired by mobile networks~\cite{MobileNetV2}.

\begin{figure*}[t]
\centering
\scriptsize
\begin{tabular}{cc}
	\figh[.3]{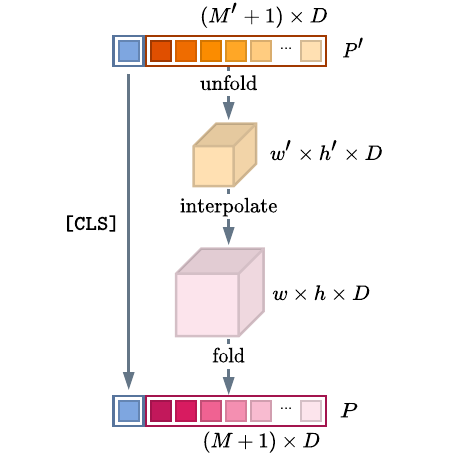} & 
	\figh[.28]{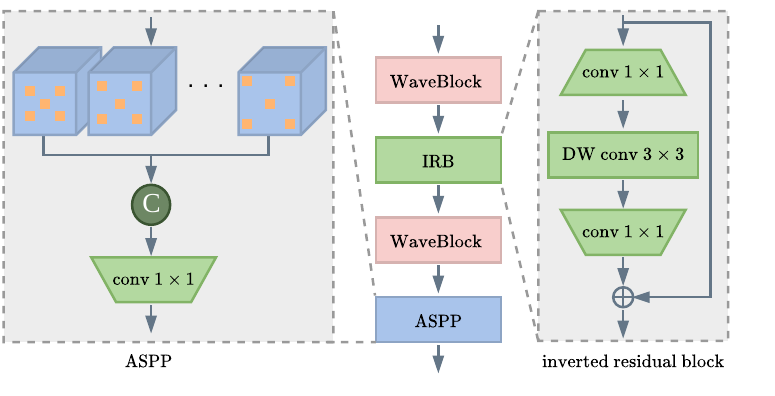} \\ 
	(a) dynamic position &
	(b) enhanced locality module \\
	embedding
\end{tabular}
\vspace{-8pt}
\caption{(a) Our \emph{dynamic position embedding} (DPE) adapts learnable position embeddings of fixed size $(M'+1) \times D$ to dynamic size $(M+1) \times D$, when using images of different size per mini-batch. (b) Our \emph{enhanced locality module} (ELM) in the local branch enhances locality of interactions and consists of an inverted residual block (IRB), an \emph{\`a trous spatial pyramid pooling} (ASPP) and two WaveBlock layers.}
\label{fig:elm-dpe}
 \vspace{-10pt}
\end{figure*}

\subsection{Detailed model}
\label{sec:model}

According to the design principles discussed above, we provide a detailed account of our full model. In our ablation study (\autoref{sec:ablation}), ideas and components are assessed individually.

\paragraph{Dynamic position embedding (DPE)}

The position embeddings of the transformer encoder (\autoref{sec:prelim}) are represented by a learnable matrix $P$ that is assumed to be of the same size as the input sequence $X$, that is, $(M+1) \times D$. When image size is different in each mini-batch, the input resolution $w \times h$ and $M = wh$ are also different. But how can $P$ change size while being learnable, that is, maintained across mini-batches?

We address this inconsistency by actually representing the position embeddings by a learnable matrix $P' = [\vp_\cls; \vp'_1; \dots; \vp'_{M'}]$ of fixed size $(M'+1) \times D$, where $M' = w'h'$ and $w' \times h'$ is some fixed spatial resolution. As shown in \autoref{fig:elm-dpe}(a), at each mini-batch, the sequence $\vp'_1, \dots, \vp'_{M'}$ corresponding to the patch tokens is unfolded to a $w' \times h' \times D$ tensor, then interpolated and resampled as $w \times h \times D$, and finally folded back to a new sequence $\vp_1, \dots, \vp_M$. 	Pre-pending $\vp_\cls$ again, which remains unaffected, yields the position embedding
\begin{equation}
	P = [\vp_\cls; \vp_1; \dots; \vp_M]
\label{eq:pos}
\end{equation}
of dynamic size $(M+1) \times D$ per mini-batch. We call this method \emph{dynamic position embedding} (DPE).


\paragraph{Multi-layer \cls/patch features}

The input sequence $X$ and the position embedding sequence $P$ are first added
\begin{equation}
	Z^0 = X + P = [\vz_\cls^0; \vz_1^0; \dots; \vz_M^0] \in \real^{(M+1) \times D}.
\label{eq:layer-0}
\end{equation}
This new sequence is the input to the transformer encoder. Let $f^\ell: \real^{(M+1) \times D} \to \real^{(M+1) \times D}$ be the mapping of layer $\ell$ of the encoder and
\begin{equation}
	Z^\ell = f^\ell(Z^{\ell-1})= [\vz_\cls^\ell; \vz_1^\ell; \dots; \vz_M^\ell] \in \real^{(M+1) \times D}
\label{eq:layer-l}
\end{equation}
be its output sequence, for $\ell=1, \dots, L$, where $L$ is the number of layers.

Given a hyper-parameter $k \in \{1, \dots, L\}$, \emph{multi-layer} \cls and \emph{patch features} are collected from the sequences $Z^{L-k+1}, \dots, Z^L$ of the last $k$ layers:
\begin{align}
	F_c &= [\vz_\cls^{L-k+1}; \dots; \vz_\cls^L] \in \real^{k \times D}
		\label{eq:multi-cls} \\
	F_p &= [A^{L-k+1}; \dots; A^L] \in \real^{k \times w \times h \times D},
		\label{eq:multi-patch}
\end{align}
where $A^\ell \in \real^{w \times h \times D}$ is the sequence $\vz_1^\ell, \dots, \vz_M^\ell$ of patch token embeddings of layer $\ell$, unfolded into a $w \times h \times D$ tensor, recalling that $M = wh$.

\begin{figure}[t]
\centering
\scriptsize
\setlength{\tabcolsep}{1pt}
\begin{tabular}{cccc}
	\fig[.24]{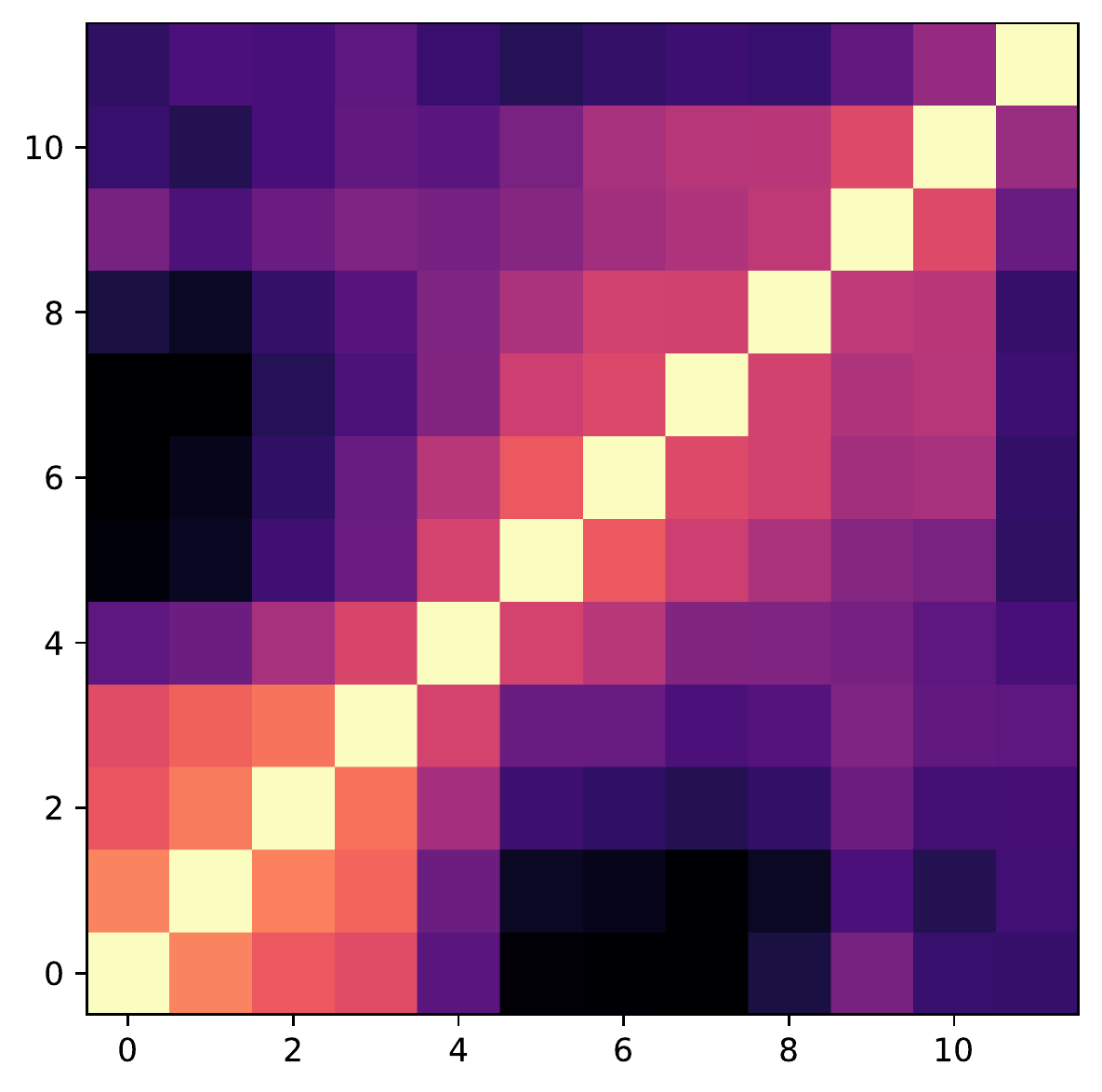} &
	\fig[.24]{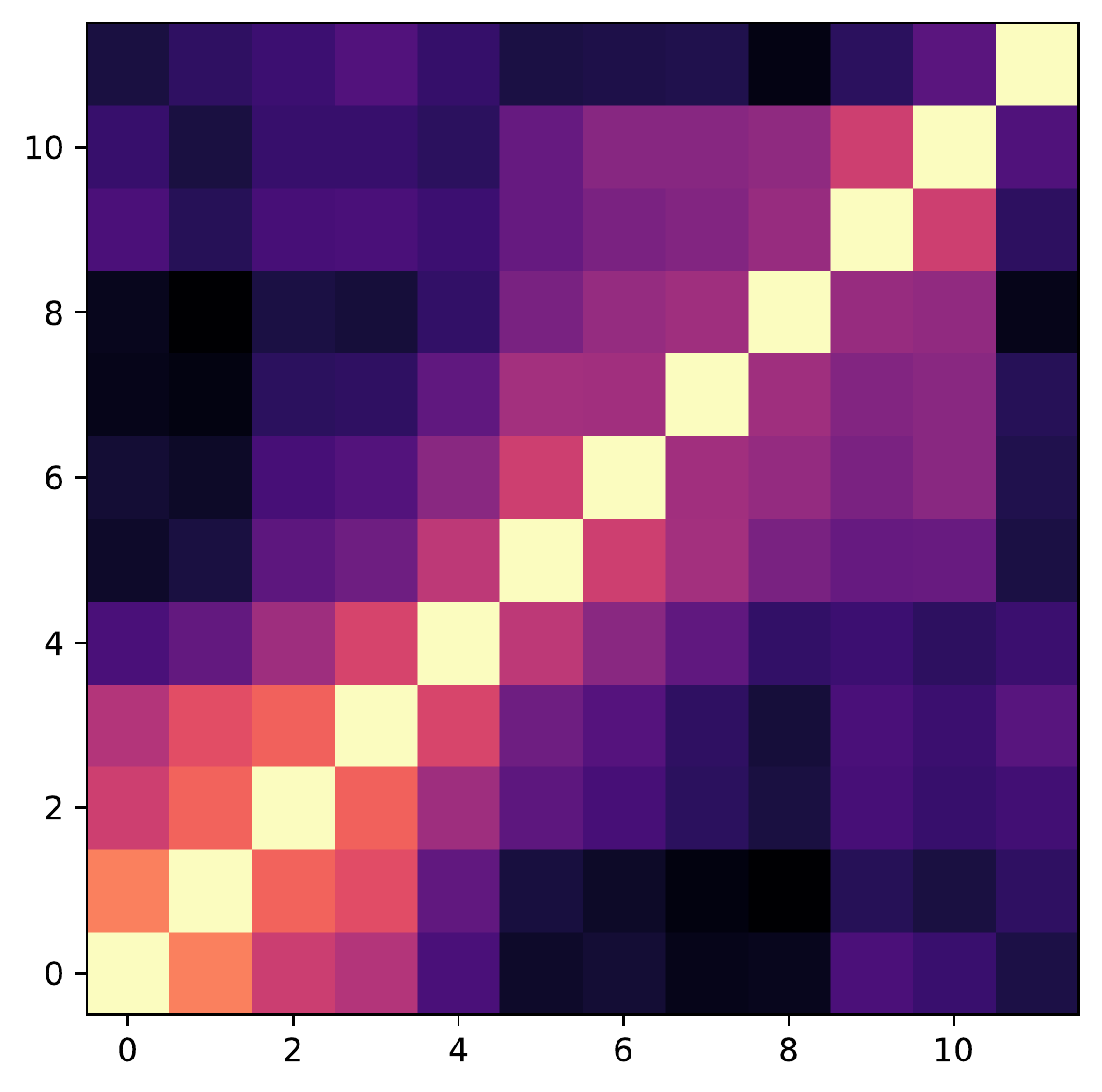} &
	\fig[.24]{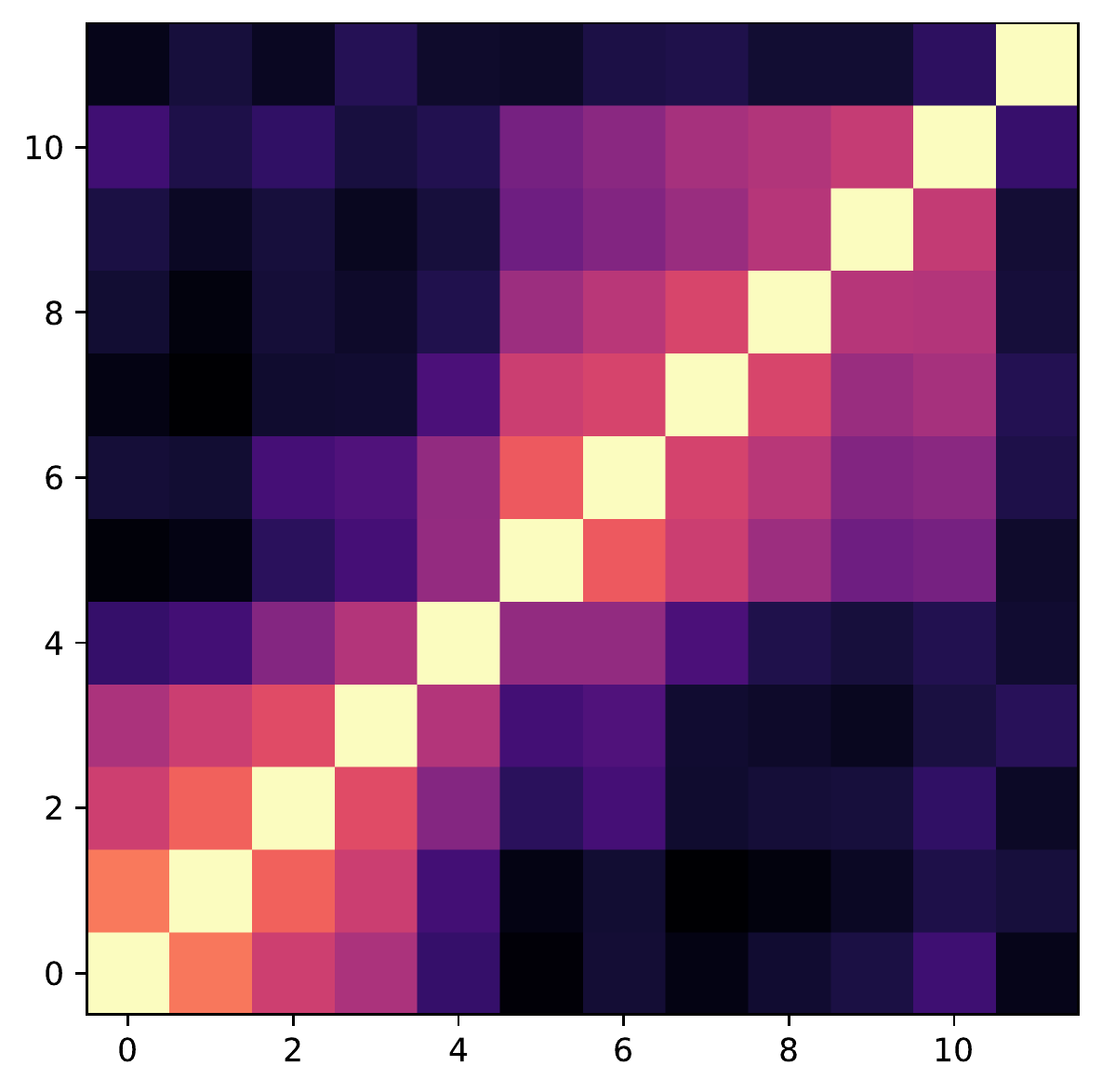} &
	\fig[.24]{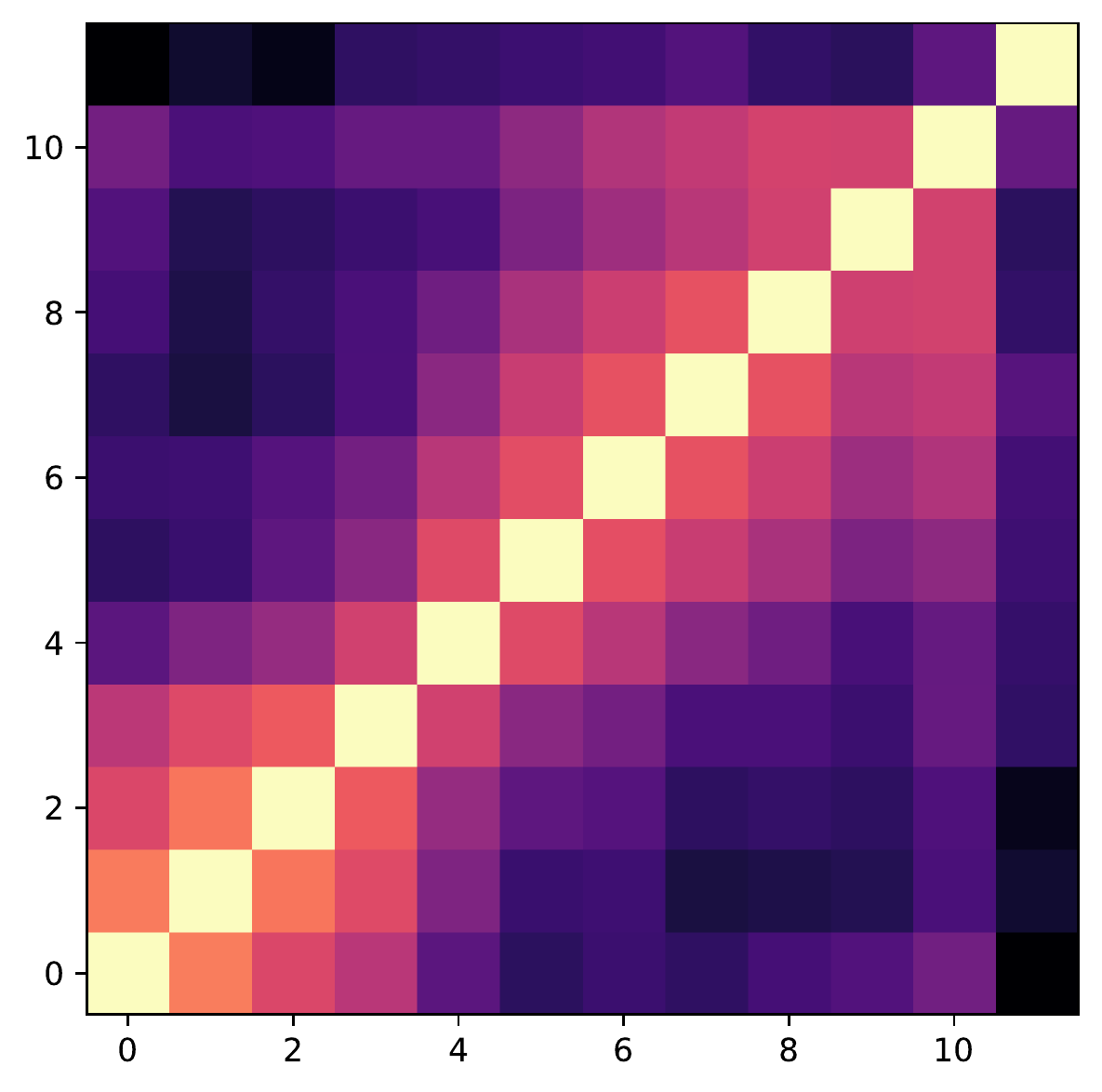} \\
	(a) NC-clean &
	(b) SfM-120k &
	(c) GLDv1-noisy &
	(d) GLDv2-clean
\end{tabular}
\vspace{-8pt}
\caption{\emph{Centered kernel alignment} (CKA)~\cite{cka, raghu2021vision} heatmap between all pairs of transformer encoding layers of our models trained on different datasets (\autoref{sec:setup}). For a given pair of layers, CKA is the correlation coefficient of the Gram matrices of the feature tensors of the two layers, averaged over the training set.}
\label{fig:cka}
\vspace{-10pt}
\end{figure}

Apart from the ablation on $k$ in \autoref{sec:ablation}, can we already get an idea whether $k > 1$ is meaningful? From \autoref{fig:cka}, the answer is yes. Layers 1-4 and 6-11 tend to group by correlation, while layer 5 is correlated with both groups. It is thus not clear which of the layers 1-11 stand out as more distinctive. What is crystal clear is that the last layer 12 is totally uncorrelated with all others.


\paragraph{Global and local branches}

The \emph{global branch}, shown above the encoder in \autoref{fig:arch}, takes as input the multi-layer \cls features $F_c$~\eq{multi-cls} and embeds them in a $N$-dimensional space
\begin{equation}
	\vu_c = \fc(F_c) \in \real^N.
\label{eq:global}
\end{equation}
using a fully connected layer (FC).
The \emph{local branch}, shown below the encoder in \autoref{fig:arch}, takes as input the multi-layer patch features $F_p$~\eq{multi-patch}, containing rich spatial information. We apply $1 \times 1$ convolution to reduce the number of channels effectively from $kD$ to $D$:
\begin{equation}
	Y = \conv_{1 \times 1}(F_p) \in \real^{w \times h \times D}
\label{eq:local-conv}
\end{equation}
Then, we apply our enhanced locality module (ELM), described below, and we fuse with $Y$, choosing from a number of alternative functions studied in the supplementary:
\begin{equation}
	Y' = \fuse(Y, \elm(Y)) \in \real^{w \times h \times D},
\label{eq:fuse}
\end{equation}
We obtain an $N$-dimensional embedding from the local branch by global average pooling (GAP) over the spatial dimensions ($w \times h$), followed by an FC layer:
\begin{equation}
    \vu_p = \fc(\gap(Y')) \in \real^N.
\label{eq:local}
\end{equation}


\paragraph{Enhanced locality module (ELM)}

As shown in \autoref{fig:elm-dpe}, our \emph{enhanced locality module} (ELM) consists of an \emph{inverted residual block} (IRB)~\cite{MobileNetV2} followed by \`a trous spatial pyramid pooling (ASPP)~\cite{Chen2017RethinkingAC}. IRB is wrapped by two WaveBlock (WB)~\cite{waveblock} layers, serving as feature-level augmentation:
\begin{equation}
    \elm(Y) = \aspp(\wav(\irb(\wav(Y)))) \in \real^{w \times h \times D}.
\label{eq:elm}
\end{equation}
IRB is a lightweight convolutional layer, where convolution is separable over the spatial and channel dimensions. In particular, it consists of a $1 \times 1$ convolution (expansion from $D$ to $D' > D$), a $3 \times 3$ depthwise convolution and another $1 \times 1$ convolution (squeeze from $D'$ to $D$) layer.
ASPP acquires multi-scale spatial context information: Feature maps obtained by \`a trous (dilated) convolution at multiple dilation rates $r_1, \dots, r_n$ are concatenated and reduced back from $nD$ to $D$ dimensions by a $1 \times 1$ convolution.


\paragraph{Image representation}

Finally, as shown on the right of \autoref{fig:arch}, we obtain a global $N$-dimensional image representation by concatenating $\vu_c$~\eq{global} with $\vu_p$~\eq{local} and applying dropout, a fully connected layer and batchnorm (BN):
\begin{equation}
    \vu = \bn(\fc(\dropout([\vu_c; \vu_p]))) \in \real^N,
\label{eq:out}
\end{equation}
reducing the dimensions from $2N$ to $N$.

%% file: tex/exp-setup.tex
\section{Experiments}
\label{sec:exp}

\subsection{Setup}
\label{sec:setup}

\paragraph{Training sets}

There are a number of open landmark datasets commonly used for training in image retrieval studies, including \emph{neural code} (NC)~\cite{Babenko01}, \emph{structure-from-motion} 120k (SfM-120k)~\cite{Radenovic01}, \emph{Google landmarks v1} (GLDv1)~\cite{delf} and \emph{v2} (GLDv2)~\cite{Weyand01}. Most of these datasets are noisy because they were obtained by text search. For example, many images contain no landmarks. Clean sets are also available where noise has has been removed in different ways~\cite{Gordo01,Radenovic01,Weyand01}. Overall, we use NC-clean, SfM-120k, GLDv1-noisy and GLDv2-clean as training sets in our experiments. More details are in the supplementary.


\paragraph{Evaluation sets/metrics}

We use Oxford5 (\oxf5k)~\cite{Philbin01}, Paris6k (\paris6k)~\cite{Philbin02}, Revised Oxford (\roxf~or \rox), and Paris (\rpar~or \rpa)~\cite{RITAC18} as evaluation sets in our experiments. We also use one million distractors (\r 1m)~\cite{RITAC18} in some experiments.
We use the Medium and Hard protocols of~\cite{RITAC18}.
Performance is measured by \emph{mean Average Precision} (mAP) and \emph{mean precision at} 10 (mP@10).


\paragraph{Architecture}

The architecture is chosen as R50+ViT-B/16~\cite{VIT} with 98M parameters, pre-trained on ImageNet-21k~\cite{imagenet21k}. That is, we use Resnet50 as CNN stem and ViT-B/16 as transformer encoder. Factor 16 is the downsampling ratio of the stem. Through all encoder layers, the embedding dimension is $D = 768$, the default of ViT-B. The additional components of DToP have only 0.3M parameters. The choice of architecture is among 15 candidates (14 vision transformers and the selected hybrid model), of which we benchmarked 12 by training on SfM-120k and measuring retrieval performance using global features, as detailed in the supplementary.


\paragraph{Implementation details}

We conduct a detailed ablation study in \autoref{sec:ablation} and in the supplementary. Default settings are as follows. The number of multi-layer features~\eq{multi-cls},\eq{multi-patch} is $k = 6$. The set of dilation rates $\{r_1, \dots, r_n\}$ of ASPP~\eq{elm} is $\{6, 12, 18\}$. The dimension of the output feature vector $\vu$ is $N = 1,536$. Training settings are detailed in the supplementary. Supervised whitening~\cite{Radenovic01} is applied except for GLDv2-clean, where the performance drops. At inference, the batch normalization of~\eq{out} is removed and we adopt a multi-scale representation using 3 scales~\cite{Gordo01,Radenovic01}. We do not consider local descriptor matching~\cite{delf, Oriane, cao2020unifying, Giorgos, RITAC18} or re-ranking~\cite{Iscen_2017_CVPR,Yang01}.

%% file: tex/exp-bench.tex
\subsection{Main results}
\label{sec:bench}

\autoref{tab:sota} compares our \Ours models against the state of the art (SOTA), grouped by representation type (global or local descriptors) and by training set.
We use a global descriptor, so local descriptors should be for reference only, but we do make some comparisons to show that a better training set or model can compensate for a larger representation. To the best of our knowledge, we are the first to organize results by training set and train a model on all commonly used training sets in the literature for fair comparison. More results are in the supplementary, including re-ranking with diffusion~\cite{Iscen_2017_CVPR}.

\begin{table*}
\input{tex/tab-sota}
\vspace{-10pt}
\caption{
mAP comparison by local/global descriptors and by training set; comparison of our \ours \vs global descriptor SOTA per training set.
Bold: best results per training set. A: AlexNet; V16: VGG16; R50/101:Resnet50/101.
By default, all competitor results are as reported by authors.
$\dagger/\ddagger$: training using official/our code;
$\exists$: evaluation using official pre-trained model;
$\square$: non-cropped queries.
}
\label{tab:sota}
\vspace{3pt}
\end{table*}

\paragraph{Comparisons with global descriptors}

Under global descriptors and all training sets, we achieve SOTA performance in almost all evaluation cases. We do not use local descriptor based re-ranking~\cite{delf, cao2020unifying, Oriane, fwtan_instance_2021} or aggregation~\cite{Giorgos,RITAC18}, or any other re-ranking such as diffusion~\cite{Iscen_2017_CVPR,Yang01}. Of course, as detailed in the supplementary, our model is not comparable to others in terms of backbone or pre-training, but it is our objective to advance vision transformers on image retrieval without introducing a new architecture.

On GLDv2-clean, comparing with published results of DOLG~\cite{yang2021dolg}, we improve by 0.7\%, 1.0\% on \rox, \rpa Medium and by 7.6\%, 7.8\% on \rox, \rpa Hard. DOLG is still better on $+$\r1m, but we have not been able to reproduce the published results using the official pre-trained model and the same problem has been encountered by the community\footnote{https://github.com/feymanpriv/DOLG-paddle/issues/3}. This may be due to having used non-cropped queries or not; the official code\footnote{https://github.com/tanzeyy/DOLG-instance-retrieval} uses non-cropped queries by default, which is against the protocol~\cite{RITAC18}. By evaluating the DOLG official pre-trained model, we outperform it on all datasets except \oxf5k. For completeness, we also report non-cropped queries for both DOLG and our \ours separately, where again we outperform DOLG on all datasets.


\paragraph{Comparisons with vision transformer studies}

Such studies are only few~\cite{gkelios2021investigating,IRT,dino,fwtan_instance_2021}. They all use global descriptors, except~\cite{fwtan_instance_2021}, which uses a transformer to re-rank features obtained by CNN rather than to encode images.
On global descriptors/SfM-120k, comparing with IRT~\cite{IRT}, we improve by 13.4\%, 10.4\% on \rox, \rpa Medium and by 14.7\%, 16.2\% on \rox, \rpa Hard.
On local descriptors/GLDv2-clean, comparing with RRT~\cite{fwtan_instance_2021}, we improve by 6.3\%, 5.3\% on \rox, \rpa Medium and by 4.3\%, 7.8\% on \rox, \rpa Hard. This shows the advantage of our model over the local descriptors of RRT.


\paragraph{Comparisons with local descriptor aggregation}

Studies using local descriptors generally work better at the cost of more memory and more complex search process. Our study is not of this type, but we do show better performance results if we allow different training sets---which has been the norm before our work. Comparing our \ours on GLDv2 with HOW~\cite{Giorgos} on SfM-120k, we improve by 2.7\%, 10.4\% on \rox, \rpa Medium and by 3.4\%, 20.5\% on \rox, \rpa Hard. On the same training set (SfM-120k), we improve on \rpa but lose on \rox. We also lose from very recent improvements~\cite{wu2021learning,superfeatures}.


\paragraph{Comparisons with local descriptor-based re-ranking}

Such methods first filter by global descriptors, then re-rank candidates by local descriptors, which is even more complex than local descriptor aggregation. On GLDv2-clean, comparing with DELG~\cite{cao2020unifying}, we improve by 0.9\%, 4.8\% on \rox, \rpa Medium and by 0.5\%, 10.1\% on \rox, \rpa Hard.

\begin{table}[t]
\centering
\scriptsize
\begin{tabular}{l*{7}{c}} \toprule
\mr{2}{\Th{Dim}} & \mr{2}{\Th{Oxf5k}} & \mr{2}{\Th{Par6k}} & \mc{2}{\Th{Medium}} & \mc{2}{\Th{Hard}} \\ \cmidrule(l){4-7}
 & & &  \rox & \rpa & \rox & \rpa \\ \midrule
DINO (ViT-S)~\cite{dino}\ &  -- & -- & 51.5 & 75.3 & 24.3  & 51.6 \\
IRT  (DeiT-B)~\cite{IRT}\ & -- & -- & 55.1 & 72.7 &  28.3  & 49.6 \\
Ours (DeiT-B) & 85.9 & 90.1 & 62.3 & 78.1 & 33.6 & 54.5 \\
Ours (ViT-B) & \tb{89.7} & \tb{92.7} & \tb{68.5} & \tb{83.1} & \tb{43.0} & \tb{65.8} \\ \bottomrule
\end{tabular}
\vspace{-9pt}
\caption{mAP comparison of our model with previous approaches using vision transformers as backbones for retrieval. Training on SfM-120k.}
\label{tab:rt}
\end{table}

\paragraph{Image retrieval using vision transformers}

In \autoref{tab:rt}, we compare with the few previous approaches using vision transformers as backbones for image retrieval, to highlight our progress. Both~\cite{IRT, dino} use global descriptors from the same transformer encoder like we do, although they use different pre-training settings. Our improvement by 10-20\% mAP is significant progress.

%% file: tex/tab-sota.tex
\newcolumntype{L}[1]{>{\raggedright\let\newline\\\arraybackslash\hspace{0pt}}m{#1}}
\newcolumntype{C}[1]{>{\centering\let\newline\\\arraybackslash\hspace{0pt}}m{#1}}
\newcolumntype{R}[1]{>{\raggedleft\let\newline\\\arraybackslash\hspace{0pt}}m{#1}}
\def\arraystretch{1.13}
\newcommand\cw{0.8cm}

\centering
\footnotesize
\setlength\extrarowheight{-1pt}
\setlength{\tabcolsep}{0pt}
\resizebox{\textwidth}{!}
{
\begin{tabular}{@{~}L{4.1cm}C{0.03cm}C{0.03cm}C{0.03cm}L{0.03cm}|C{\cw}C{0.9cm}|C{\cw}C{\cw}C{\cw}C{\cw}C{\cw}C{\cw}C{\cw}C{\cw}|C{\cw}C{\cw}C{\cw}C{\cw}C{\cw}C{\cw}C{\cw}C{\cw}}
	\toprule
	\multicolumn{5}{@{~}c|}{\multirow{3}{*}{\Th{Method}}} &
	\multicolumn{2}{c|}{\Th{Base}}  & \multicolumn{8}{c|}{\Th{Medium}} & \multicolumn{8}{c}{\Th{Hard}} \\
	\cmidrule{6-23}

	& & & & &  {\fontsize{7}{6}\selectfont \oxf5k} & {\fontsize{7}{6}\selectfont \paris6k} & \multicolumn{2}{c}{{\fontsize{7}{6}\selectfont \rox}} & \multicolumn{2}{c}{{\fontsize{7}{6}\selectfont \rox+\r1m}} & \multicolumn{2}{c}{{\fontsize{7}{6}\selectfont \rpa}} & \multicolumn{2}{c|}{{\fontsize{7}{6}\selectfont \rpa+\r1m}} & \multicolumn{2}{c}{{\fontsize{7}{6}\selectfont \rox}} & \multicolumn{2}{c}{{\fontsize{7}{6}\selectfont \rox+\r1m}} & \multicolumn{2}{c}{{\fontsize{7}{6}\selectfont \rpa}} & \multicolumn{2}{c}{{\fontsize{7}{6}\selectfont \rpa+\r1m}} \\

	& & & & & \tiny mAP &  \tiny mAP & \tiny mAP & \tiny mP@10 & \tiny mAP & \tiny mP@10 & \tiny mAP & \tiny mP@10 & \tiny mAP & \tiny mP@10 & \tiny mAP & \tiny mP@10 & \tiny mAP & \tiny mP@10 & \tiny mAP & \tiny mP@10 & \tiny mAP & \tiny mP@10 \\
	\midrule

	\rowcolor{lightgray}
	\multicolumn{23}{c}{\Th{Local Descriptors (NC-clean)}} \\ \midrule
	DELF-DIR-R101~\cite{delf} & & & & &  90.0 & {95.7} & -- & -- & -- & --  &  -- & -- & -- & -- & -- & -- & -- & -- & -- & -- & -- & -- \\
	\midrule

	\rowcolor{lightgray}
	\multicolumn{23}{c}{\Th{Local Descriptors (SfM-120k)}} \\ \midrule
	DELF-ASMK-SP~\cite{RITAC18}  & & & & &  -- & -- & 67.8 & \tb{87.9} & -- & -- & 76.9 & \tb{99.3} & -- & -- & 43.1 & \tb{62.4} & -- & -- & 55.4 & {93.4} & -- & --  \\
	DSM-MAC-R101~\cite{Oriane}  & & & & & -- & -- & 62.7 & 83.7 & 44.4 & 72.3 &  75.7 & 98.7  & 50.4 & 96.4 & 35.4 & 51.6 & 20.6 & 32.3 &  53.1 & 88.6 & 22.7 & 72.1  \\
	DSM-GeM-R101~\cite{Oriane}  & & & & & -- & -- & 65.3 & 87.1 & 47.6 & \tb{76.4} &  77.4 & 99.1  & 52.8 & \tb{96.7} & 39.2 & 55.3 & 23.2 & \tb{37.9} & 56.2 & \tb{89.9}  & 25.0 & \tb{74.6}  \\
	HOW-ASMK-R50~\cite{Giorgos}  & & & & &  -- & -- & 79.4 & -- & -- & -- & 81.6 & --  & -- & -- &  56.9 & -- & -- & -- & 62.4 & -- & -- & -- \\
	FIRe-R50~\cite{superfeatures}  & & & & &  -- & -- & \tb{81.8} & -- & 66.5 & -- & \tb{85.3} & --  & \tb{67.6} & -- &  61.2 & -- & 40.1 & -- & \tb{70.0} & -- & \tb{42.9} & -- \\
	MDA-R50~\cite{wu2021learning}  & & & & &  -- & -- & \tb{81.8} & -- & \tb{68.7} & -- &  83.3 & --  & 64.7 & -- &  \tb{62.2} & -- & \tb{45.3} & -- &  66.2 & -- & 38.9 & -- \\
	\midrule

	\rowcolor{lightgray}
	\multicolumn{23}{c}{\Th{Local Descriptors (GLDv2-clean)}} \\ \midrule
	DELG-GeM-R50~\cite{cao2020unifying}  & & & & &  -- & -- & 78.3 & -- &  67.2 & -- & 85.7 & -- & 69.6 & -- & 57.9 & -- & 43.6 & -- & 71.0 & -- &  45.7 & --  \\
	DELG-GeM-R101~\cite{cao2020unifying} & & & & &  -- & -- & \tb{81.2} & -- & \tb{69.1} & -- & \tb{87.2} & --  & \tb{71.5} & -- & \tb{64.0} & -- & \tb{47.5} & -- & 72.8 & -- & 48.7 & --  \\
	RRT-R50~\cite{fwtan_instance_2021} & & & & &  -- & -- & 78.1 &  -- &  67.0 &  -- & 86.7 &  -- & 69.8 &  -- & 60.2 &  -- & 44.1 &  -- & \tb{75.1} &  -- & \tb{49.4} & --  \\
	\midrule

	\rowcolor{lightgray}
	\multicolumn{23}{c}{\Th{Global Descriptors (Off-the-shelf, pre-trained on ImageNet)}} \\ \midrule
	SPoC-V16~\cite{Babenko03,RITAC18}                 & & & & & 53.1 & -- & 38.0 & 54.6 & 17.1 & 33.3 & 59.8 & 93.0 & 30.3 & 83.0 & 11.4 & 20.9 & 0.9 & 2.9 & 32.4 & 69.7 & 7.6 & 30.6 \\
	SPoC-R101~\cite{RITAC18}                           & & & & & -- & -- & 39.8 & 61.0 & 21.5 & 40.4 & 69.2 & 96.7 & 41.6 & 92.0 & 12.4 & 23.8 & 2.8 & 5.6 & 44.7 & 78.0 & 15.3 & 54.4 \\
	MAC-V16~\cite{Radenovi01,RITAC18}         & & & & & 80.0 & 82.9 & 37.8 & 57.8  & 21.8  & 39.7 & 59.2 & 93.3  & 33.6 & 87.1 & 14.6  & 27.0  & \tb{7.4}  & 11.9 & 35.9  & 78.4  & 13.2 & 54.7 \\
	MAC-R101~\cite{RITAC18}                    & & & & & -- & -- & 41.7 & 65.0 & 24.2 & 43.7 & 66.2 & 96.4 & 40.8 & 93.0 & 18.0 & \tb{32.9} & 5.7 & \tb{14.4} & 44.1 & 86.3 & 18.2 & \tb{67.7} \\
	RMAC-V16~\cite{Radenovi01,RITAC18}         & & & & & \tb{80.1} & 85.0 & 42.5 & 62.8 & 21.7 & 40.3 & 66.2 & 95.4 & 39.9 & 88.9 & 12.0 & 26.1 & 1.7 & 5.8 & 40.9 & 77.1 & 14.8 & 54.0 \\
	RMAC-R101~\cite{RITAC18}                   & & & & & -- & -- & \tb{49.8} & \tb{68.9} &  \tb{29.2} &  \tb{48.9} &  \tb{74.0} &  \tb{97.7} &  \tb{49.3} &  \tb{93.7} &  \tb{18.5} & 32.2 &  4.5 &  13.0 &  \tb{52.1} &  \tb{87.1} &  \tb{21.3} &  67.4 \\
	DINO-R50~\cite{dino}                               & & & & &  -- & -- &  35.4 & -- & -- & -- & 55.9  & -- & -- & -- & 11.1   & -- & -- & -- & 27.5 & -- & -- & --  \\
	DINO-ViT-S~\cite{dino}      & & & & &  -- & -- &  41.8 & -- & -- & -- & 63.1  & -- & -- & -- & 13.7   & -- & -- & -- & 34.4  & -- & -- & --  \\
	ViT-B~\cite{gkelios2021investigating} & & & & &  64.7 & \tb{87.8} &  -- & -- & -- & -- & --  & -- & -- & -- & -- & -- & -- & -- & -- & -- & -- & --  \\
	\midrule

	\rowcolor{lightgray}
	\multicolumn{23}{c}{\Th{Global Descriptors (NC-clean)}} \\ \midrule
	RMAC-R101~\cite{Gordo01,RITAC18}   & & & & & 86.1 & \tb{94.5} & 60.9 & 78.1 & 39.3 & 62.1 & 78.9 & \tb{96.9} & \tb{54.8} & \tb{93.9} & 32.4 & 50.0 & 12.5 & 24.9 & 59.4 & 86.1 & 28.0 & \tb{70.0} \\
	GLAM-R101~\cite{SCH01}     & & & & &  77.8 & 85.8 & 51.6 & -- & -- & -- & 68.1 & -- & -- & -- & 20.9 & -- & -- & -- & 44.7 & -- & -- & -- \\
	DOLG-R101~\cite{yang2021dolg}$^\dagger$ & & & & & 73.1 & 79.5 & 47.0 & 67.1 & -- & -- & 61.5 & 92.1 & -- & -- & 20.0 & 33.7 & -- & -- & 33.1 & 68.3 & -- & -- \\
	\tb{\ours-R50+ViT-B} (ours)  & & & & & \tb{86.2} & 92.9 &  \tb{64.8} & \tb{81.3} & \tb{46.5} & \tb{64.3} & \tb{82.0} & 96.4 & 54.6 & 92.9 & \tb{41.3} &  \tb{56.9} & \tb{21.5} & \tb{36.6} & \tb{63.2} &  \tb{87.4} & \tb{28.9} & 66.9  \\
	\midrule

	\rowcolor{lightgray}
	\multicolumn{23}{c}{\Th{Global Descriptors (GLDv1-noisy)}} \\ \midrule
	SOLAR-R101~\cite{Ng01}            & & & & & -- & -- & 69.9 & \tb{86.7} & 53.5 & \tb{76.7} & 81.6  & \tb{97.1} & 59.2 & 94.9 & 47.9 & 63.0 & 29.9 & 48.9 &  64.5 & \tb{93.0} & 33.4 & 81.6 \\
	GeM-R101~\cite{Weyand01}    & & & & & -- & -- & 68.9 & -- & -- & -- & 83.4 & --  & -- & -- & 45.3  & -- &  -- & -- & 67.2 & -- & -- & --  \\
	GLAM-R101~\cite{SCH01}     & & & & & \tb{92.8} & 95.0 & \tb{73.7} & -- &  -- & -- &
	83.5 & -- &  -- & -- & 49.8 & -- &  -- & -- & 69.4 & --  & -- & -- \\
	DINO-DeiT-S~\cite{dino}      & & & & &  -- & -- &  51.5 & -- & -- & -- & 75.3  & -- & -- & -- & 24.3  & -- & -- & -- & 51.6  & -- & -- & --  \\
	\tb{\ours-R50+ViT-B} (ours) & & & & &  88.8 & \tb{95.7} & 71.6 & 85.1 & \tb{55.8} & 75.9 & \tb{88.7} & 96.6 & \tb{67.8} & \tb{95.6} & \tb{51.7} & \tb{65.1} & \tb{30.9} & \tb{49.3} & \tb{75.9} &  90.4 & \tb{44.1} & \tb{82.0}  \\
	\midrule

	\rowcolor{lightgray}
	\multicolumn{23}{c}{\Th{Global Descriptors (SfM-120k)}} \\ \midrule
	RMAC-R101~\cite{Gordo01}$^\ddagger$   & & & & & 79.0 & 86.3 & 53.5 & 76.9& -- & -- & 68.3 & 97.7 & -- & -- & 25.5 & 42.0 & -- & --& 42.4 & 83.6 & -- & -- \\
	GeM-A~\cite{Radenovic01,RITAC18}      & & & & & 67.7 & 75.5 &  43.3  & 62.1 & 24.2 & 42.8 & 58.0  & 91.6 & 29.9 & 84.6 & 17.1 & 26.2 & 9.4 & 11.9  & 29.7 & 67.6 & 8.4 & 39.6 \\
    GeM-V16~\cite{Radenovic01,RITAC18}    & & & & & 87.9 & 87.7 &  61.9  & 82.7 & 42.6 & 68.1 & 69.3 & 97.9 & 45.4 & 94.1 & 33.7 & 51.0 &  19.0 & 29.4 & 44.3 & 83.7 & 19.1 & 64.9 \\
	GeM-R101~\cite{Radenovic01,RITAC18}   & & & & & 87.8 & \tb{92.7} & 64.7 & 84.7 & 45.2 & \tb{71.7} & 77.2 & \tb{98.1} & 52.3 & \tb{95.3} & 38.5 & 53.0 & 19.9 & 34.9 & 56.3 & \tb{89.1} & 24.7 & \tb{73.3} \\
	AGeM-R101~\cite{gu2018attention}   & & & & &  -- & -- & 67.0 & -- & -- & -- & 78.1 & --  & -- & -- & 40.7 & -- & -- & -- & 57.3 & --  & -- & --  \\
	SOLAR-R101~\cite{Ng01}$^\dagger$       & & & & & 78.5 & 86.3 & 52.5 & 73.6 & -- & -- & 70.9 & \tb{98.1} & -- & -- & 27.1 & 41.4 & -- & -- & 46.7 & 83.6 & -- & -- \\
	GeM-R101~\cite{Weyand01}$^\dagger$  & & & & & 79.0 & 82.6 & 54.0 & 72.5 & -- & -- & 64.3 & 92.6 & -- & -- & 25.8 & 42.2 & -- & -- & 36.6 & 67.6 & -- & -- \\
	GLAM-R101~\cite{SCH01}$^\ddagger$  & & & & & \tb{89.7} & 91.1 & 66.2 & -- & -- & -- & 77.5 & -- & -- & -- & 39.5 & -- & -- & -- & 54.3 & -- & -- & -- \\
	DOLG-R101~\cite{yang2021dolg}$^\dagger$ & & & & & 72.8 & 74.5 & 46.4 & 66.8 & -- & -- & 56.6 & 91.1 & -- & -- & 18.1 & 27.9 & -- & -- & 26.6 & 62.6 & -- & -- \\
	IRT-DeiT-B~\cite{IRT}       & & & & &  -- & -- &  55.1 & -- & -- & -- & 72.7  & -- & -- & -- & 28.3  & -- & -- & -- & 49.6  & -- & -- & --  \\
	\tb{\ours-R50+ViT-B} (ours)  & & & & & \tb{89.7} & \tb{92.7} &  \tb{68.5} & \tb{85.4} & \tb{48.9} & \tb{71.7} & \tb{83.1} & 96.4 & \tb{56.5} & 94.0 & \tb{43.0} & \tb{56.9} & \tb{24.7} & \tb{38.9} & \tb{65.8} & \tb{89.1} & \tb{30.3} & 69.6 \\
	\midrule

	\rowcolor{lightgray}
	\multicolumn{23}{c}{\Th{Global Descriptors (GLDv2-clean)}} \\ \midrule
	GeM-R101~\cite{Weyand01}   & & & & & -- & -- & 76.2 & -- & -- & -- & 87.3 & --  & -- & -- &  55.6  & -- & -- & -- & 74.2 & -- & -- & -- \\
	GLAM-R101~\cite{SCH01}     & & & & & 94.2 & 95.6 & 78.6 & 88.2 & 68.0 & 82.4 & 88.5 & \tb{97.0} & 73.5 & {94.9} & 60.2 & 72.9 & 43.5 & 62.1 & 76.8 & 93.4 & 53.1 & 84.0   \\
	DELG-GeM-R50~\cite{cao2020unifying}   & & & & & -- & -- & 73.6 & -- & 60.6 & -- & 85.7 & -- & 68.6 & &  51.0 & -- & 32.7 & -- &   71.5 & -- &  44.4 & --   \\
	DELG-GeM-R101~\cite{cao2020unifying}   & & & & & -- & -- & 76.3 & -- & 63.7 & -- & 86.6 & -- & 70.6 & & 55.6 & -- & 37.5 & -- &  72.4 & -- &  46.9 & --   \\
	DOLG-R50~\cite{yang2021dolg}    & & & & & -- & -- & 80.5 & -- & 76.6 & -- & 89.8  & -- & 80.8 & -- &  58.8 & -- & 52.2 & -- & 77.7 & -- & 62.8 & --   \\	
	DOLG-R101~\cite{yang2021dolg}  & & & & & -- & -- & 81.5 & -- & \tb{77.4} & -- & 91.0  & -- &  \tb{83.3} & -- & 61.1 & -- & \tb{54.8} & -- & 80.3 & --  & \tb{66.7} & -- \\
	DOLG-R101~\cite{yang2021dolg}$^\exists$      & & & & & \tb{95.5} & 95.9 & 78.8 & 91.6 & 64.2 & 82.1 & 87.8  & 96.6 & 68.7 & 94.1 &  58.0 & 74.8 & 37.3 & 57.7 & 74.1 & 91.1 & 45.1 & 80.0   \\
	\tb{\ours-R50+ViT-B} (ours) & & & & & 95.0 & \tb{97.0} & \tb{82.1} & \tb{91.7} & 70.9 & \tb{83.9} & \tb{92.0} & 96.6 & 81.9 & \tb{96.4} & \tb{64.5} & \tb{77.4} &  49.0 & \tb{66.6} & \tb{82.9}  & \tb{94.3}  & 64.0 & \tb{90.6} \\
	\midrule

	DOLG-R101~\cite{yang2021dolg}$^{\exists \square}$  & & & & & 91.7 & 96.2 & 79.3 & 93.2 & 71.3 & 89.1 & 89.2  & 98.9 & 74.7 & 97.7 & 57.2 & 73.0 & 43.4 & 62.6 & 76.6 & 94.1 & 53.6 & 89.7   \\
	\tb{\ours-R50+ViT-B} (ours)$^\square$ & & & & &  94.2 & 97.6 & {84.4} &  {94.1} &  {78.9} &  {91.3} & {92.3} & {97.1} & {85.4} &  96.9 & {64.8} & {76.7} & {57.1} &  {72.1} & {84.6} & {95.4} & {71.2} & {94.6}  \\
	\bottomrule
\end{tabular}
}

%% file: tex/exp-vis.tex
\subsection{Visualization}
\label{sec:vis}

\begin{figure*}[t]
\centering
	\fig[1.0]{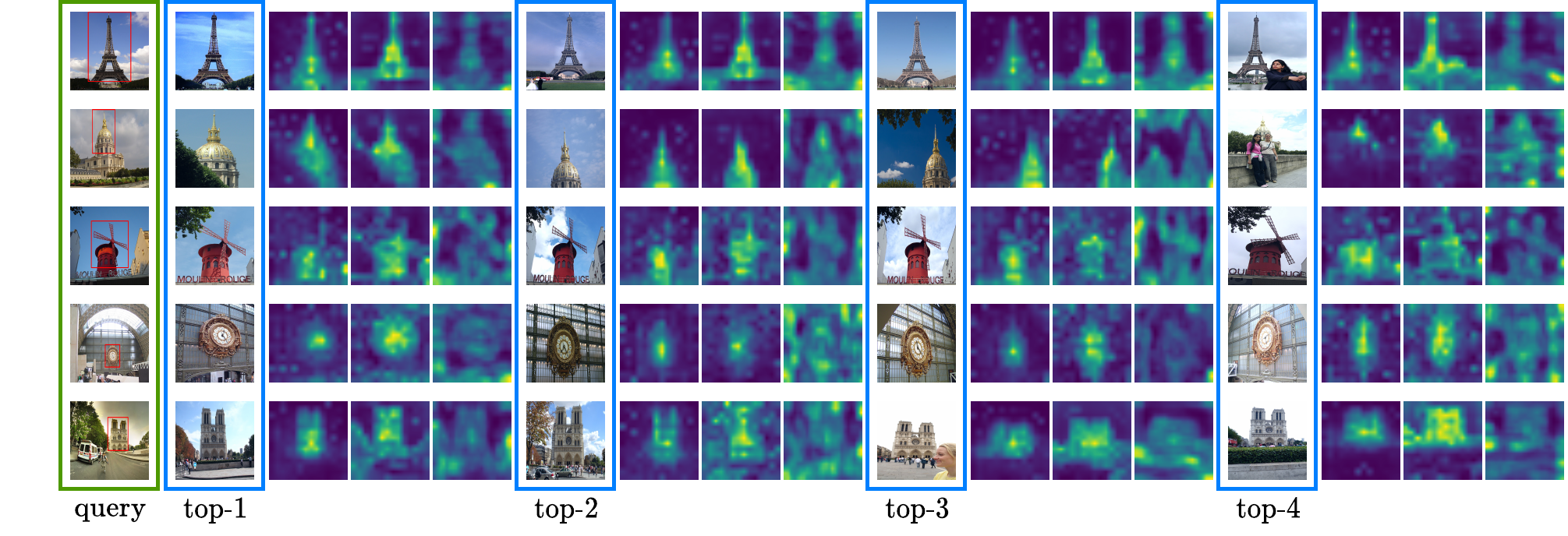}
\vspace{-10pt}
\caption{Examples of ranking by our model and spatial attention. On the left (green box) is the query, followed by 4 top-ranking results, including images (blue box) attention maps by our \ours (full model), our baseline model (\autoref{tab:design}) and Resnet101. For our transformer model (full and baseline), we show the attention map between the \cls and all the other tokens. For Resnet we do the same, using GAP to obtain a vector that plays the role of the \cls token embedding.}
\label{fig:vis}
\end{figure*}

\paragraph{Ranking and spatial attention}

\autoref{fig:vis} shows examples, for a number of queries, of the top-ranking images by our \ours, along with attention maps. Our model is attending only the object of interest, not the background. Whereas, Resnet101 is also attending the background to a great extent. Our full model also attends more clearly the foreground than the baseline.
More visualizations are given in the supplementary, including t-SNE embeddings.

%% file: tex/exp-ablation.tex
\subsection{Ablation study}
\label{sec:ablation}

Unless otherwise stated, all ablation experiments are conducted on SfM-120k. Additional experiments are in the supplementary, including experiments on 13 different transformer backbones, the number of layers $k$, DPE against baselines, ELM components, the feature dimension $N$, mini-batch sampling, multi-scale features at inference as well as fusion alternatives for~\eq{fuse}.

\paragraph{Design/algorithmic ablation}

We assess the effect of different choices and components to the performance of our model. Our baseline is the plain ViT-B/16 transformer using raw patch tokens, with a plain \cls token representation from the last layer ($F_c = \vz_\cls^L$), mapped to $N$ dimensions by an FC layer~\eq{global}. The baseline is trained with group-size sampling~\cite{Yokoo2020TwostageDR} and our DPE~\eq{pos}; this setting is ablated separately. Adding the CNN stem amounts to switching to R50+ViT-B/16 hybrid model. Adding the local and global branches includes multi-layer \cls~\eq{multi-cls} and patch~\eq{multi-patch} features respectively, where the local branch replaces the plain \cls token representation ($F_c = \vz_\cls^L$) by~\eq{multi-cls} when present. In~\eq{out}, only $\vu_c$~\eq{global}, $\vu_p$~\eq{local} is present when only the global, local branch is present respectively. Removing ELM amounts to setting $Y' = Y$ in~\eq{fuse}.

\begin{table}[ht]
\centering
\scriptsize
\setlength{\tabcolsep}{1.8pt}
\begin{tabular}{l*{12}{c}} \toprule
& \Th{CNN} & \Th{global} & \Th{local} & & \mr{2}{\Th{ELM}} & \mr{2}{\Th{Oxf5k}} & \mr{2}{\Th{Par6k}} & \mc{2}{\Th{Medium}} & \mc{2}{\Th{Hard}} \\ \cmidrule(l){9-12}
& \Th{Stem} & \Th{branch} & \Th{branch} & & & & &\rox & \rpa & \rox & \rpa \\
\midrule
& &     &     &     &     & 77.7  & 85.9 & 52.6 & 76.0 & 26.6 & 52.0 \\
& & \ch &     &     &     & 76.6  & 87.3 & 54.7 & 77.0 & 27.7 & 54.8 \\
& & \ch & \ch &     &     & 78.3  & 89.7 & 57.9 & 78.2 & 24.2 & 54.4 \\
& & \ch & \ch &     & \ch & 81.5  & 89.8 & 61.4 & 79.7 & 32.5 & 57.4 \\ 
\midrule
& \ch &  &     &  &     &  81.2  & 86.4 & 55.5 & 76.2 & 31.4 & 52.1 \\
&\ch & \ch &     &  &     & 88.3  & 91.9 & 66.6 & \tb{83.6} & 41.9 & \tb{67.8} \\ 
&  \ch& \ch & \ch & &     & \tb{89.8}  & 91.2 & 67.6 & 81.1 & 40.7 & 62.5 \\
& \ch & \ch & \ch &  & \ch & 89.7 & \tb{92.7} & \tb{68.5} & 83.1 & \tb{43.0} & {65.8} \\
\bottomrule
\end{tabular}
\vspace{-8pt}
\caption{mAP comparison of variants of our model with/without different components. Training on SfM-120k. ELM: enhanced locality module~\eq{elm}.}
\label{tab:design}
\end{table}

As shown in \autoref{tab:design}, the greatest improvement comes from the CNN stem when combined with the global branch, confirming the importance of the inductive bias of convolution in the early layers and the complementarity of the \cls features of the last layers. In the absence of the CNN stem, each component (global/local branch, ELM) is improving the performance. By contrast, when the CNN stem is present, the contribution of the local branch and ELM is inconsistent across \rox and \rpa. These two components can thus be thought as lightweight alternatives to the CNN stem, improving locality in the last layers.

%% file: tex/supp.tex
\clearpage





\appendix



\section{More experiments}

\subsection{Setup}

\paragraph{Network size and pretraining}

The ViT-B transformer encoder and the R50+ViT-B hybrid have 86M and 98M parameters respectively and are pretrained on ImageNet-21k. The additional components of DToP have only 0.3M parameters. The common competitors Resnet50/101 have 25M and 44M parameters respectively and are pretrained on ImageNet-1k. Despite its lower dimensionality (1,536 \vs 2,048 dimensions for Resnet101), our DToP-R50+ViT-B is of course in a priviledged position over Resnet101 in terms of both network size and pretraining data. Still, it is important that a transformer reaches SOTA on image retrieval for the first time. Our objective is not to introduce a new architecture.


\paragraph{Training settings}

We apply random cropping, illumination and scaling augmentation. We use a batch size of 64 and the ArcFace loss with margin 0.15. We optimize by stochastic gradient descent with momentum 0.9, initial learning rate $10^{-3}$ and cosine learning rate decay with the decay factor $10^{-4}$. We apply warm-up of 0, 3, 5 and 5 epochs on NC-clean, SfM-120k, GLDv1-noisy and GLDv2-clean, respectively. We implement our method on PyTorch and we train our models on 8 TITAN RTX 3090Ti GPUs.

\subsection{Benchmarking of vision transformer models}

Vision transformer studies have exploded in a short period of time, but very few concern image retrieval. We perform for the first time an extensive empirical study to benchmark a large number of vision transformer models on image retrieval and choose the best performing one as our default backbone.

\paragraph{Candidate models}

We fine-tune on image retrieval training sets, so we only consider models that are pre-trained on ImageNet-1k or ImageNet-21k~\cite{imagenet21k}. In particular, we consider the models shown in \autoref{tab:study}.

As global image representation, all models use a \cls (classification) token embedding, while PiT~\cite{pit} and DeiT~\cite{deit} also provide a distillation token embedding. As a local image representation, patch token embeddings can be used for all models, while for the hybrid R50+ViT-B, features of the CNN stem can also be used. Certain pre-trained models, in particular Swin~\cite{Swin}, ConViT~\cite{convit} and TNT~\cite{tnt}, cannot handle multi-scale input. These models are not compatible with the group-size sampling approach that we adopt for training~\cite{Yokoo2020TwostageDR}. This constraint is not due to the architecture itself but to the way the code is written, and although it would be certainly possible to fix, this would require some effort. We therefore exclude them from our benchmark.

\begin{table}
\centering
\scriptsize
\setlength{\tabcolsep}{2.7pt}
\begin{tabular}{lccccc} \toprule
	\mr{2}{\Th{Model}} & \mr{2}{\Th{CLS}} & \mr{2}{\Th{Dist}} & \mr{2}{\Th{Patch}} & \mr{2}{\Th{CNN}} & \mr{2}{\Th{MS}} \\ \\ \midrule
	Swin~\cite{Swin}                     & \ch &     & \ch &     &     \\
	ConViT~\cite{convit}                 & \ch &     & \ch &     &     \\
	TNT~\cite{tnt}                       & \ch &     & \ch &     &     \\
	ViL~\cite{vil}                       & \ch &     & \ch &     & \ch \\
	CvT~\cite{cvt}                       & \ch &     & \ch &     & \ch \\
	LocalViT~\cite{localvit}             & \ch &     & \ch &     & \ch \\
	Patch~\cite{patchvisiontransformer}  & \ch &     & \ch &     & \ch \\
	T2T~\cite{tokens}                    & \ch &     & \ch &     & \ch \\
	DeepViT~\cite{dvit}                  & \ch &     & \ch &     & \ch \\
	LV-ViT~\cite{tokenlabeling}                   & \ch &     & \ch &     & \ch \\
	PiT~\cite{pit}                       & \ch & \ch & \ch &     & \ch \\
	DeiT (DeiT-B)~\cite{deit}            & \ch & \ch & \ch &     & \ch \\
	ViT (ViT-B)~\cite{VIT}               & \ch &     & \ch &     & \ch \\
	ViT (R50+ViT-B)                      & \ch &     & \ch & \ch & \ch \\ \bottomrule
\end{tabular}
\caption{Feature types that can be extracted from different pre-trained vision transformer models considered in our study. CLS: classification token; Dist: distillation token; Patch: patch tokens. CNN: convolutional stem (hybrid model). MS: can handle multi-scale input at training (required in our experiments).}
\label{tab:study}
\vspace{5pt}
\end{table}


\paragraph{Setup}

We take all models as pre-trained on either ImageNet-1k or ImageNet-21k and fine-tune them on SfM-120k~\cite{Radenovic01}. We use only the global branch~\eq{global} on multi-layer \cls features~\eq{multi-cls} or the local branch~\eq{local} on multi-layer patch features~\eq{multi-patch}. In the former case, we also evaluate multi-layer \emph{distillation} features, replacing \cls by the distillation token where it exists, \ie, PiT~\cite{pit} and DeiT~\cite{deit}. In the latter case, we do not use the enhanced locality module (ELM), that is, we set $Y' = Y$ in~\eq{fuse}. At the output, instead of~\eq{out}, we only use an FC layer with output feature dimension $N = 768$. We use group-size sampling~\cite{Yokoo2020TwostageDR} with our dynamic position embedding (DPE)~\eq{pos}. We use default training settings, except without warm-up.

At inference, we evaluate each model with exactly the same type of features as at training (\cls, distillation or patch), applying supervised whitening~\cite{Radenovic01} on multi-scale features~\cite{Radenovic01} and measuring mAP on the evaluation sets.

\begin{table*}[t]
\centering
\footnotesize
\setlength{\tabcolsep}{3pt}
\resizebox{\textwidth}{!}
{
\begin{tabular}{lcc|ccccccc|cccccc} \toprule
\mr{4}{\Th{Model}} &
\mr{4}{\Th{Pre-Train}} &
\mc{1}{\mr{4}{\begin{tabular}{c}\Th{Params} \\ (M)\end{tabular}}} &
\mc{7}{\Th{Global (CLS/Dist)}} &
\mc{6}{\Th{Local (Patch)}} \\ \cmidrule{4-16}
& & \mc{1}{} &
\mr{2}{\Th{Token}} &
\mr{2}{\Th{Oxf5k}} &
\mr{2}{\Th{Par6k}} &
\mc{2}{\Th{Medium}}   &
\mc{2}{\Th{Hard}}  &
\mr{2}{\Th{Oxf5k}} &
\mr{2}{\Th{Par6k}} &
\mc{2}{\Th{Medium}}  &
\mc{2}{\Th{Hard}} \\ \cmidrule{7-10} \cmidrule{13-16}
& & \mc{1}{} & & & & {\fontsize{5}{4}\selectfont \rox} &  {\fontsize{5}{4}\selectfont \rpa} & {\fontsize{5}{4}\selectfont \rox} & \mc{1}{\fontsize{5}{4}\selectfont \rpa} &   &  &  {\fontsize{5}{4}\selectfont \rox} &  {\fontsize{5}{4}\selectfont \rpa} &  {\fontsize{5}{4}\selectfont \rox} & {\fontsize{5}{4}\selectfont \rpa} \\ \midrule
T2T-ViT-24~\cite{tokens} & ImageNet-1k & 64 & \textsc{CLS} & 35.3 &  49.3 &  14.2 & 35.5 & 2.0 & 10.6 & {65.0} & {72.0} & {42.0} & {52.1} & {18.6} & {22.8} \\
DeepViT-B~\cite{dvit} & ImageNet-1k & 48 & \textsc{CLS} &  43.9 &  56.9 & 21.1 & 41.7 & 4.2 & 13.4 & 50.4 & 62.1 & {27.7} & 46.2 & {7.1} & 18.2 \\
LocalViT-S~\cite{localvit} & ImageNet-1k & 22 & \textsc{CLS} & 65.0 & 71.3 & 38.4 & 53.0 & 14.9 & 22.1 & 62.1 & 71.4 & 36.5 & 50.9 & 11.2 & 19.1 \\
LV-ViT-M~\cite{tokenlabeling} & ImageNet-1k & 39 & \textsc{CLS} & {72.4} & {79.2} & {45.4} & {65.2} & {19.6} & 28.9 &  42.1 & 57.1 & 25.6 & 45.8 & 11.4 & 20.9 \\ \midrule
Patch-ViT-B~\cite{patchvisiontransformer} & ImageNet-21k & 86 & \textsc{CLS} & 28.8 &  38.7 & 13.5 & 28.5 & 1.7 & 8.0 & {59.5} & {68.9} & {34.5} & {50.6} & {13.1} & {18.6} \\
ViL-B~\cite{vil} & ImageNet-21k & 55 & \textsc{CLS} & 42.6  & 56.0 & 20.7 & 39.5 &  3.1 & 12.8 & 43.6 & 51.1 & 20.1 & 36.4 & 3.1 & 10.8 \\
CvT-21~\cite{cvt} & ImageNet-21k & 31 & \textsc{CLS} &{84.0}  & {87.8} & {61.3} & {78.8} & {30.8} & {56.7} & 77.6 & 81.6 & 54.6 & \tb{76.4} & 25.7 & \tb{52.9} \\ \midrule
\mr{2}{PiT-B~\cite{pit}} & \mr{2}{ImageNet-1k} & \mr{2}{73} & \textsc{CLS} &  69.0 & 77.0 & 43.6 &  59.1 & 19.8 & 28.7 & \mr{2}{70.9} & \mr{2}{81.2} & \mr{2}{42.4} &  \mr{2}{64.8} & \mr{2}{18.9} & \mr{2}{35.5} \\
& & & \textsc{Dist} &  66.1 & 78.5 & 41.5 & 60.6 & 15.5 & 30.1 \\ \midrule
\mr{2}{DeiT-B~\cite{deit}} & \mr{2}{ImageNet-1k} & \mr{2}{86} & \textsc{CLS} & 82.3  & 84.1 & 59.6 & 67.6 & 29.1 & 46.5 &  \mr{2}{\tb{80.7}} & \mr{2}{78.5} & \mr{2}{\tb{54.7}} & \mr{2}{64.5} & \mr{2}{22.7} & \mr{2}{37.2} \\
& &  & \textsc{Dist} & {84.2} & {85.5} & {62.4} & 71.2 & {32.6} & {49.7} \\ \midrule
ViT-B~\cite{VIT} & ImageNet-21k & 86 & \textsc{CLS} & {76.2}  & {83.4} & {49.3} & 70.9 & {19.6} & 46.4 & 60.4 & 81.0 & 38.6 & 69.2 & 12.0 & 46.9 \\
R50+ViT-B~\cite{VIT} & ImageNet-21k & 98 & \textsc{CLS} & \tb{84.3} & \tb{87.9} & \tb{62.6} & \tb{79.6} & \tb{37.9} & \tb{64.8} & 74.6 & \tb{82.8} & 50.9 & 69.4 & \tb{26.9} & 46.5 \\ \bottomrule
\end{tabular}
}
\caption{mAP comparison of different pre-trained vision transformer models, using multilayer \cls or distillation (\textsc{Dist}) token features from our global branch~\eq{global}, or multi-layer patch features from the local branch~\eq{local}, without ELM~\eq{elm}. All options give rise to a global representation of $N = 768$ dimensions; patch features undergo global average pooling. Fine-tuning on SfM-120k~\cite{Radenovic01} using default settings.}
\label{tab:bench}
\vspace{5pt}
\end{table*}

\paragraph{Results}

\autoref{tab:bench} shows the results of the benchmark. In the majority of cases, we can see that \cls outperforms patch features: CvT-21~\cite{cvt}, LocalViT-S~\cite{localvit}, LV-ViT-M~\cite{tokenlabeling}, DeiT-B~\cite{deit}, ViT-B~\cite{VIT}, R50+ViT-B~\cite{VIT}. However, in many cases, the opposite holds: Patch-ViT-B~\cite{patchvisiontransformer}, T2T-ViT-24~\cite{tokens}, DeepViT-B~\cite{dvit}. In few cases, the performance is similar or inconsistent: ViL-B~\cite{vil}, PiT-B~\cite{pit}. As for the distillation token, it works consistently better than \cls for DeiT-B~\cite{deit}, but for PiT-B~\cite{pit} there is no clear winner. Overall, we observe that when multi-layer features are used for image retrieval, \cls does not work necessarily better than global average pooling. ImageNet-21k is also not necessarily better than the smaller ImageNet-1k as a training set; for example, Patch-ViT-B~\cite{patchvisiontransformer} using \cls performs worse overall using this training set.

Using the \cls token, the hybrid model R50+ViT-B~\cite{VIT} is a clear winner overall, also better than the same model using patch features. The second and third best are CvT-21~\cite{cvt} using \cls and DeiT-B~\cite{deit} using distillation token, respectively. The hybrid model has more parameters (98M) than the plain ViT-B~\cite{VIT}/DeiT-B~\cite{deit} (86M) and a lot more than CvT-21~\cite{cvt} (31M) and the majority of models. It is also pretrained on the larger ImageNet-21k training set. In this sense, this is not a fair comparison. There are many other factors that are not shown here, like distillation from a stronger model by DeiT-B~\cite{deit} and PiT-B~\cite{pit}.

We choose R50+ViT-B~\cite{VIT} as the default backbone in our experiments for two reasons: (a) Our objective is to explore how much vision transformers can improve on image retrieval; and (b) its improvement over other models on the Hard protocol is more pronounced, implying it is a much stronger model. We suspect that its improvement is also significant in the presence of the challenging \r1m distractors, although we cannot benchmark all models for this. What we can suggest as a more lightweight alternative is CvT-21~\cite{cvt}.

\begin{table}[t]
\scriptsize
\centering
\setlength{\tabcolsep}{2.5pt}
\begin{tabular}{lcc|cccc} \toprule
\mr{2}{\Th{Train Set}}           & \mc{2}{(a) \Th{Statistics}}     & \mc{4}{(b) \Th{Improvement}}                                              \\ \cmidrule{2-7}
                                 & \Th{\#Classes} & \Th{\#Images}  & \mc{2}{\Th{Previous SOTA}}                 & \Th{Ours} & \Th{Gain}        \\ \midrule
NC-noisy~\cite{Babenko01}        & 672            &  213,678       &                                &           &           &                  \\
\tb{NC-clean}~\cite{Gordo01}     & 581            & 27,965         & RMAC~\cite{Gordo01,RITAC18}    & 57.9      & 62.8      & \gain{+4.9}      \\
\tb{SfM-120k}~\cite{Radenovic01} & 713            & 117,369        & GeM~\cite{Radenovic01,RITAC18} & 59.2      & 65.1      & \gain{+5.9}      \\
\tb{GLDv1-noisy}~\cite{delf}     & 14,951         & 1,225,029      & SOLAR~\cite{Ng01}              & 66.0      & 72.0      & \gain{+6.0}      \\
{GLDv2-noisy}~\cite{Weyand01}    & 203,094        & 4,132,914      &                                &           &           &                  \\
\tb{GLDv2-clean}~\cite{Weyand01} & 81,313         & 1,580,470      & DOLG~\cite{yang2021dolg}       & 78.0      & 80.4      & \gain{+2.4}      \\ \bottomrule
\end{tabular}
\vspace{3pt}
\caption{(a) Open landmark training set statistics. Bold: datasets used in our experiments. (b) Average mAP comparison of our \vs previous SOTA models based on global features. Average taken over \rox and \rpa evaluation sets in $\mathcal{R}$Medium and $\mathcal{R}$Hard protocols (four columns) in \autoref{tab:sota}.}

\label{tab:data-prog}
\end{table}

\begin{figure}[t]
\small
\centering
\setlength{\tabcolsep}{0pt}
\pgfplotsset{
	dataset/.style={
            width=.545\linewidth,
		height=.55\linewidth,
		font=\tiny,
		enlarge x limits=.2,
		x tick label style={rotate=20,anchor=east},
	}
}
\begin{tabular}{cc}
\hspace{-12pt}
\pgfplotstableread{
	id          classes   images
	NC-clean       581     27965
	SfM-120k       713    117369
	GLDv1-noisy  14951   1225029
	GLDv2-clean  81313   1580470
}{\data}
\extfig{data}{
\begin{tikzpicture}[%
	img/.style={purple,mark=*},
]
\begin{axis}[%
	dataset,
	xlabel={Training set},
	ylabel={\# classes},
	legend pos=north west,
	axis y line*=right,
	ymin=0,ymax=120000,
	ybar,bar width=20pt,
	symbolic x coords={NC-clean,SfM-120k,GLDv1-noisy,GLDv2-clean},
]
	\addplot[orange,fill=orange!50,nodes near coords,every node near coord/.style={black},] table[x=id,y=classes] \data; \leg{\# classes};
	\addplot[mark=none,draw=none,sharp plot,line legend,legend image post style=img]  table[x=id,y=images]  \data; \leg{\# images};
\end{axis}
\begin{axis}[%
	dataset,
	ylabel={\# images},
	grid=none,
	axis x line=none,
	axis y line*=left,
	ymin=0,
	symbolic x coords={NC-clean,SfM-120k,GLDv1-noisy,GLDv2-clean},
]
	\addplot[img]  table[x=id,y=images]  \data;
\end{axis}
\end{tikzpicture}
}
&
\pgfplotstableread{
	id           prev  ours
	NC-clean     57.9  62.8
	SfM-120k     59.2  65.1
	GLDv1-noisy  66.0  72.0
	GLDv2-clean  78.0  80.4
}{\data}
\extfig{sota}{
\begin{tikzpicture}[%
]
\begin{axis}[%
	dataset,
	xlabel={Training set},
	ylabel={mAP},
	legend pos=north west,
	ybar,bar width=8pt,nodes near coords,
	symbolic x coords={NC-clean,SfM-120k,GLDv1-noisy,GLDv2-clean},
]
	\addplot[blue,fill=blue!50,] table[x=id,y=prev] \data; \leg{Previous SOTA};
	\addplot[red, fill=red!50,]  table[x=id,y=ours] \data; \leg{Our models};
\end{axis}
\end{tikzpicture}
}
\\
(a) Training set statistcs &
(b) Our \vs previous models \\
\end{tabular}
\caption{(a) Open landmark training set statistics, from \autoref{tab:data-prog}(a). (b) Average mAP comparison of our \vs previous SOTA models based on global features. Average taken over \rox and \rpa test sets in $\mathcal{R}$Medium and $\mathcal{R}$Hard protocols (four columns) in \autoref{tab:sota}. Previous SOTA models are as shown in \autoref{tab:data-prog}(b).}
\label{fig:data-prog}
\end{figure}
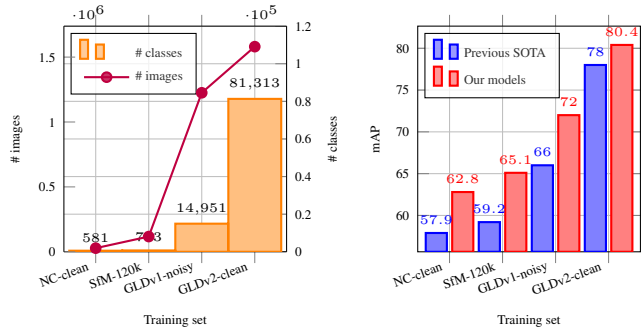

\subsection{More results}
\label{sec:more-bench}

\paragraph{Summary of progress per dataset}

\autoref{tab:data-prog}(a) and \autoref{fig:data-prog}(a) show statistics of open datasets that have been used as training sets for landmark image retrieval; in particular, number of classes and number of images per dataset. There is a variety of dataset sizes, both in terms of classes and images. In our experiments, we focus on the most commonly used datasets in the literature, that is, \emph{neural code} (NC) clean~\cite{Gordo01}, \emph{structure-from-motion} 120k (SfM-120k)~\cite{Radenovic01}, \emph{Google landmarks v1} (GLDv1) noisy~\cite{delf} and \emph{Google landmarks v2} (GLDv2) clean~\cite{Weyand01}.

\autoref{tab:data-prog}(b) and \autoref{fig:data-prog}(b) show the progress over the SOTA that we bring per dataset, based on global features. We compare separately per training set, in terms of mAP averaged over four columns of \autoref{tab:sota}. We observe that as the number of training images increases, the performance also increases. Our models bring clear improvement on all training sets, with the improvement being more pronounced on small and noisy training sets.


\subsection{More ablation experiments}
\label{sec:more-ablation}

Like \autoref{sec:ablation}, all experiments here are conducted on SfM-120k by default. We investigate the effect of different factors on the performance of our full model \ours-R50+ViT-B, including the number of layers $k$, DPE against baselines, ELM components, the feature dimension $N$, multi-scale features at inference, mini-batch sampling as well as fusion alternatives for~\eq{fuse}.

\begin{table}[t]
\centering
\scriptsize
\setlength{\tabcolsep}{3pt}
\begin{tabular}{r*{7}{c}} \toprule
\mr{3}{\begin{tabular}{c} \Th{Dim} \\ $N$ \end{tabular}} &
\mr{2}{\Th{Oxf5k}} & \mr{2}{\Th{Par6k}} &
\mc{2}{\Th{Medium}} & \mc{2}{\Th{Hard}} \\ \cmidrule{4-7}
& & & \rox & \rpa & \rox & \rpa \\ \midrule
128 & 80.7 & 89.1 & 56.7 & 78.5 & 29.6 & 59.3 \\
256 & 87.4 & 90.1 & 63.1 & 80.2 & 35.8 & 62.2 \\
512 & 88.0 & 91.1 & 64.9 & 82.1 & 38.6 & 64.7 \\
768 & 88.8 & 91.8 & 67.5 & 82.1 & 41.7 & 64.8 \\
1,024 & 86.3 &  92.7 & 64.4 & \tb{83.1} & 38.4 & \tb{66.5} \\
1,536 & \tb{89.7}  & 92.7 & \tb{68.5} & \tb{83.1} & \tb{43.0} & 65.8 \\
2,048 & 89.3 & \tb{93.0} &  65.8 & 82.9&  39.0 & 65.8 \\ \bottomrule
\end{tabular}
\caption{mAP comparison of different dimension $N$ of output features~\eq{out} of our full model. Training on SfM-120k. Using supervised whitening~\cite{Radenovic01}.}
\label{tab:dim}
\end{table}

\begin{table}[t]
\vspace{5pt}
\centering
\scriptsize
\begin{tabular}{c*{6}{c}} \toprule
\Th{Layers} &  \mr{2}{\Th{Oxf5k}} & \mr{2}{\Th{Par6k}} & \mc{2}{\Th{Medium}} & \mc{2}{\Th{Hard}} \\ \cmidrule(l){4-7}
$k$ & & & \rox & \rpa & \rox & \rpa \\ \midrule
1 & 87.2 & 92.4 & 64.9 & 81.3 & 37.6 & 62.7 \\
3 & 88.0 & 91.1 & 64.9 & 82.1 & 38.6 & 64.7 \\
6 & \tb{89.7}  & 92.7 & \tb{68.5} & \tb{83.1} & \tb{43.0} & \tb{65.8} \\
9 & 87.1 & \tb{93.1} & 66.8 & 82.4 & 41.4 & 64.2 \\
12 & 89.0 & 92.4 & 68.1 & 83.0 & 43.2 & 65.5 \\
\bottomrule
\end{tabular}
\vspace{-8pt}
\caption{mAP comparison of using different number of layers $k$ in the multi-layer features~\eq{multi-cls},\eq{multi-patch}. Training on SfM-120k.}
\label{tab:layers}
\end{table}

\paragraph{Number of layers}

\autoref{tab:layers} shows the effect of the number of layers $k$ in the multi-layer features~\eq{multi-cls},\eq{multi-patch}. The optimal number of layers is $k = 6$. The effect of $k$ is significant, especially in the Hard protocol, yielding an improvement of 6.1\% mAP over the baseline $k = 1$ on \rox. What is less understood is that $k = 12$ works as well.

\begin{table}[t]
\centering
\scriptsize
\setlength{\tabcolsep}{2.5pt}
\begin{tabular}{l*{6}{c}} \toprule
\mr{2}{\Th{PE Type}} &  \mr{2}{\Th{Oxf5k}} & \mr{2}{\Th{Par6k}} & \mc{2}{\Th{Medium}} & \mc{2}{\Th{Hard}} \\ \cmidrule(l){4-7}
 & & & \rox & \rpa & \rox & \rpa \\ \midrule
no PE & 82.8 & 85.7 & 59.7 & 73.9 & 32.5 & 47.4\\
CPE~\cite{cpe} & 85.9 & 88.8 & 62.6 & 77.9 & 37.1 & 58.2 \\
DPE (bi-cubic) & 87.6 & 91.0 & 65.2 & 82.2 & 38.3 & 64.6 \\
DPE (bi-linear) & \tb{89.7}  & \tb{92.7} & \tb{68.5} & \tb{83.1} & \tb{43.0} & \tb{65.8} \\ \bottomrule
\end{tabular}
\vspace{-9pt}
\caption{mAP comparison of our (bi-linear and bi-cubic) \emph{dynamic position embedding} (DPE)~\eq{pos} with no position embedding and with conditional position embedding (CPE)~\cite{cpe}. Training on SfM-120k.}
\label{tab:dpe}
\end{table}

\paragraph{Dynamic position embedding}

In \autoref{tab:dpe} we compare our DPE with baselines and assess the effect of interpolation type we use in DPE. Bi-linear interpolation is best, outperforming the baseline by nearly 20\% on \rpa Hard.

\begin{table}[t]
\centering
\scriptsize
\setlength{\tabcolsep}{2.5pt}
\begin{tabular}{l*{11}{c}} \toprule
& \mr{2}{CNN Stem}  & \mr{2}{IRB} & \mr{2}{ASPP} & \mr{2}{WB} & \mr{2}{\Th{Oxf5k}} & \mr{2}{\Th{Par6k}} & \mc{2}{\Th{Medium}} & \mc{2}{\Th{Hard}} \\ \cmidrule(l){8-11}
&&     & & & &  & \rox & \rpa & \rox & \rpa  \\ \midrule
&&     & \ch & \ch &  78.6 & 87.8 & 55.2 & 77.4 & 27.1 & 55.1 \\
&& \ch &     & \ch &  75.8 & 87.8 & 52.9 & 77.5 & 28.7 & 53.7 \\
&& \ch & \ch &     &  80.1 & 88.2 & 59.8 & 77.7 & 28.8 & 54.2 \\
&& \ch & \ch & \ch &  81.5  & 89.8 & 61.4 & 79.7 & 32.5 & 57.4 \\ \midrule
&\ch &  &  \ch & \ch & 84.9 & 92.7 & 64.8 & \tb{83.4} &  42.4  & 65.7 \\
&\ch & \ch &     & \ch & 85.3 & \tb{93.0} & 65.7 & 83.0 &  42.8  & 65.4  \\
&\ch & \ch & \ch &     & 87.0 & 92.0 & 66.5 & 82.7 &  42.8  & 65.1 \\
&\ch & \ch & \ch & \ch & \tb{89.7} & 92.7 & \tb{68.5} & 83.1 & \tb{43.0} & \tb{65.8} \\ \bottomrule
\end{tabular}
\vspace{-9pt}
\caption{mAP comparison of variants of enhanced locality module (ELM) with/without different components. IRB: inverted residual block~\cite{MobileNetV2}; ASPP: \`a trous spatial pyramid pooling~\cite{Chen2017RethinkingAC}; WB: WaveBlock~\cite{waveblock}. Training on SfM-120k.}
\label{tab:elm}
\end{table}

\paragraph{ELM components}

In \autoref{tab:elm} we study the effect of different components of the enhanced locality module (ELM). It is clear that all components contribute to the performance of ELM in the absence of the CNN stem. In the hybrid architecture, they are effective on Oxf5k and \rox but not on Par6k and \rpa. This result is in agreement with \autoref{tab:design}, where ELM is most effective in the absence of other forms of inductive bias.


\paragraph{Feature dimension}

The global representation obtained by our \ours model is given by~\eq{out} and is a vector of dimension $N$. \autoref{tab:dim} shows the effect of the choice of this dimension. Clearly, the best performance is obtained by $N = 1,536$, by a larger margin on \roxf (medium or hard). A larger dimension does not necessarily mean better performance.

\begin{table}[t]
\centering
\scriptsize
\setlength{\tabcolsep}{3pt}
\begin{tabular}{l*{7}{c}} \toprule
\mr{2}{\Th{Query}} & \mr{2}{\Th{Database}} & \mr{2}{\Th{Oxf5k}} & \mr{2}{\Th{Par6k}} & \mc{2}{\Th{Medium}} & \mc{2}{\Th{Hard}} \\ \cmidrule(l){5-8}
 & & & & \rox & \rpa & \rox & \rpa \\ \midrule
Single & Single & 89.4 & 92.2 & 68.2 & 82.2 & 43.0 & 64.0\\
Multi & Single & \tb{90.1} & 92.5 & 68.6 & 82.8 & 43.1 & 65.1\\
Single & Multi &  89.4 & 92.4 & \tb{68.8} & 82.6 & \tb{43.8} & 64.9 \\
Multi & Multi &  89.7 & \tb{92.7} & 68.5 & \tb{83.1} & 43.0 & \tb{65.8} \\ \bottomrule
\end{tabular}
\caption{mAP comparison of multi-scale \vs single-scale representation on queries or database. Training on SfM-120k.}
\label{tab:scale}
\end{table}

\paragraph{Multi-scale}

Following previous work~\cite{Gordo01,Radenovic01}, we use a multi-scale image representation with 3 scales at inference by computing the output features~\eq{out} for each scale of the input image and averaging the features over scales. \autoref{tab:scale} shows the effect of using multi-scale \vs single-scale representation on queries or database. Clearly, using a multi-scale representation on the database works best. As for the queries, the results are not consistent across datasets, but the gain brought by multi-scale queries on \rpar is more than the loss on \roxf. We thus choose a multi-scale representation on both queries and database.

\begin{table}[t]
\centering
\scriptsize
\setlength{\tabcolsep}{3pt}
\begin{tabular}{l*{6}{c}} \toprule
\mr{2}{\Th{Sampling}} &  \mr{2}{\Th{Oxf5k}} & \mr{2}{\Th{Par6k}} & \mc{2}{\Th{Medium}} & \mc{2}{\Th{Hard}} \\ \cmidrule(l){4-7}
 & & & \rox & \rpa & \rox & \rpa \\ \midrule
Fixed-size & 83.2 & 90.6 & 60.5 & 79.5 & 35.7 & 59.8 \\
Group-size~\cite{Yokoo2020TwostageDR} &  \tb{89.7}  & \tb{92.7} & \tb{68.5} & \tb{83.1} & \tb{43.0} & \tb{65.8} \\ \bottomrule
\end{tabular}
\caption{mAP comparison of fixed-size ($384 \times 384$) \vs group-size sampling~\cite{Yokoo2020TwostageDR} of mini-batches at training. Training on SfM-120k.}
\label{tab:sampling}
\end{table}

\paragraph{Mini-batch sampling}

As discussed in \autoref{sec:design}, we use group-size sampling~\cite{Yokoo2020TwostageDR} to account for different sizes and aspect ratios of input images, while maintaining the same size for all images in a mini-batch. This strategy results in a dynamic image size per mini-batch and we use our dynamic position embedding~\eq{pos} in this case. \autoref{tab:sampling} shows the performance of this strategy compared with fixed-size ($384 \times 384$) images for all mini-batches. It is clear that group-size sampling improves performance by a large margin, up to 8\% on \roxf and up to 6\% on \rpar.

\begin{table}[t]
\centering
\scriptsize
\setlength{\tabcolsep}{3pt}
\begin{tabular}{l*{7}{c}} \toprule
\mr{2}{\Th{Method}} & \mr{2}{\Th{Oxf5k}} & \mr{2}{\Th{Par6k}} & \mc{2}{\Th{Medium}} & \mc{2}{\Th{Hard}} \\ \cmidrule(l){4-7}
 & & & \rox & \rpa & \rox & \rpa \\ \midrule
No fusion (w/o ELM)              & 89.8 & 91.2 & 67.6 & 81.1 & 40.7 & 62.5 \\
No fusion (w/ ELM)               & 85.9 & 91.9 & 64.6 & 82.6 & 40.6 & 65.4 \\ \midrule
Sum                              & 89.5 & 91.7 & 68.4 & 82.1 & 43.4 & 64.2 \\
Hadamard product                 & \tb{89.8} & 92.0 & 68.2 & \tb{83.1} & 43.5 & 66.0 \\
Concatenation                    & 88.8 &  92.5 &  67.5 & 82.7  &  \tb{43.9} & 64.9 \\
Fast normalized~\cite{bifpn}     & \tb{89.8} & 92.1 & \tb{68.7} & 82.4 & \tb{43.9} & 65.0 \\
Orthogonal~\cite{yang2021dolg}   & 89.7 & \tb{92.7} & 68.5 & \tb{83.1} & 43.0 & \tb{65.8} \\ \bottomrule
\end{tabular}
\caption{mAP comparison of different feature fusion methods for the input and output of ELM~\eq{fuse}. Training on SfM-120k.}
\label{tab:fuse}
\end{table}

\paragraph{Feature fusion for ELM}

In the local branch, the patch features $Y \in \real^{w \times h \times D}$~\eq{local-conv} are fused in~\eq{fuse} with the output of the \emph{enhanced locality module} (ELM), say, $U = \elm(Y) \in \real^{w \times h \times D}$. This happens because the input still has the valuable spatial information. Denoting function $\fuse$ by $h$ for brevity, Eq.~\eq{fuse} is written as
\begin{equation}
	Y' = h(Y, U).
\label{eq:fuse-2}
\end{equation}
Here we consider a number of alternatives for $h$:
\begin{align}
	\textrm{No fusion (w/o ELM)           }: \quad & h(Y, U) = Y \label{eq:fuse-no-elm} \\
	\textrm{No fusion (w/ ELM)            }: \quad & h(Y, U) = U \label{eq:fuse-none} \\
	\textrm{Sum                           }: \quad & h(Y, U) = Y + U \label{eq:fuse-add} \\
	\textrm{Hadamard product              }: \quad & h(Y, U) = Y \odot U \label{eq:fuse-hadamard} \\
	\textrm{Concatenation                 }: \quad & h(Y, U) = [Y; U] \label{eq:fuse-concat} \\
	\textrm{Fast normalized~\cite{bifpn}  }: \quad & h(Y, U) =
		\frac{w_1 Y + w_2 U}{w_1 + w_2 + \epsilon}
		\label{eq:fuse-bifpn} \\
	\textrm{Orthogonal~\cite{yang2021dolg}}: \quad & h(\vy_i, \vu_i) =
		[\vy_i - \prj_{\vu_i}(\vy_i); \vu_i]
		\label{eq:fuse-ortho}
\end{align}

Fast normalized fusion is a fusion strategy investigated as part of BiFPN~\cite{bifpn}, where $h(Y, U)$ is a linear combination of $Y, U$ with $w_1 = \relu(v_1)$, $w_2 = \relu(v_2)$ and $v_1, v_2$ are learnable parameters. It has similar effect to normalizing $v_1, v_2$ by softmax but is more efficient.

Orthogonal fusion is similar to DOLG~\cite{yang2021dolg} but differs in that we fuse two 3D tensors while DOLG fuses a 3D tensor with a vector. By representing $Y$ by a folded sequence of token embeddings $\vy_1, \dots, \vy_M \in \real^D$ and similarly $U$ by $\vu_1, \dots, \vu_M \in \real^D$, we define $h(\vy_i, \vu_i)$ per token as the Gram-Schmidt orthogonalization of vectors $\vy_i, \vu_i$, where
\begin{equation}
	\prj_{\vu}(\vy) = \frac{\inner{\vy, \vu}}{\inner{\vu, \vu}} \vu
\label{eq:gram}
\end{equation}
is the orthogonal projection of $\vy$ onto the line spanned by vector $\vu$.

\autoref{tab:fuse} shows a comparison of two non-fusion and the five fusion methods. We first observe that the output $U$ of ELM alone can be worse than the input $Y$, while the five fusion methods mostly outperform the non-fusion variants. Hence, $Y$ and $U$ contain complementary information and fusion is beneficial. Of the fusion methods, the Hadamard product, fast normalized and orthogonal are the most effective, but there is no clear winner and differences are small. We choose orthogonal fusion as default.

\begin{figure}
\scriptsize
\centering
\setlength{\tabcolsep}{0pt}
\begin{tabular}{cccc}
	 \raisebox{37pt}{Easy}                   &
	 \fig[.3]{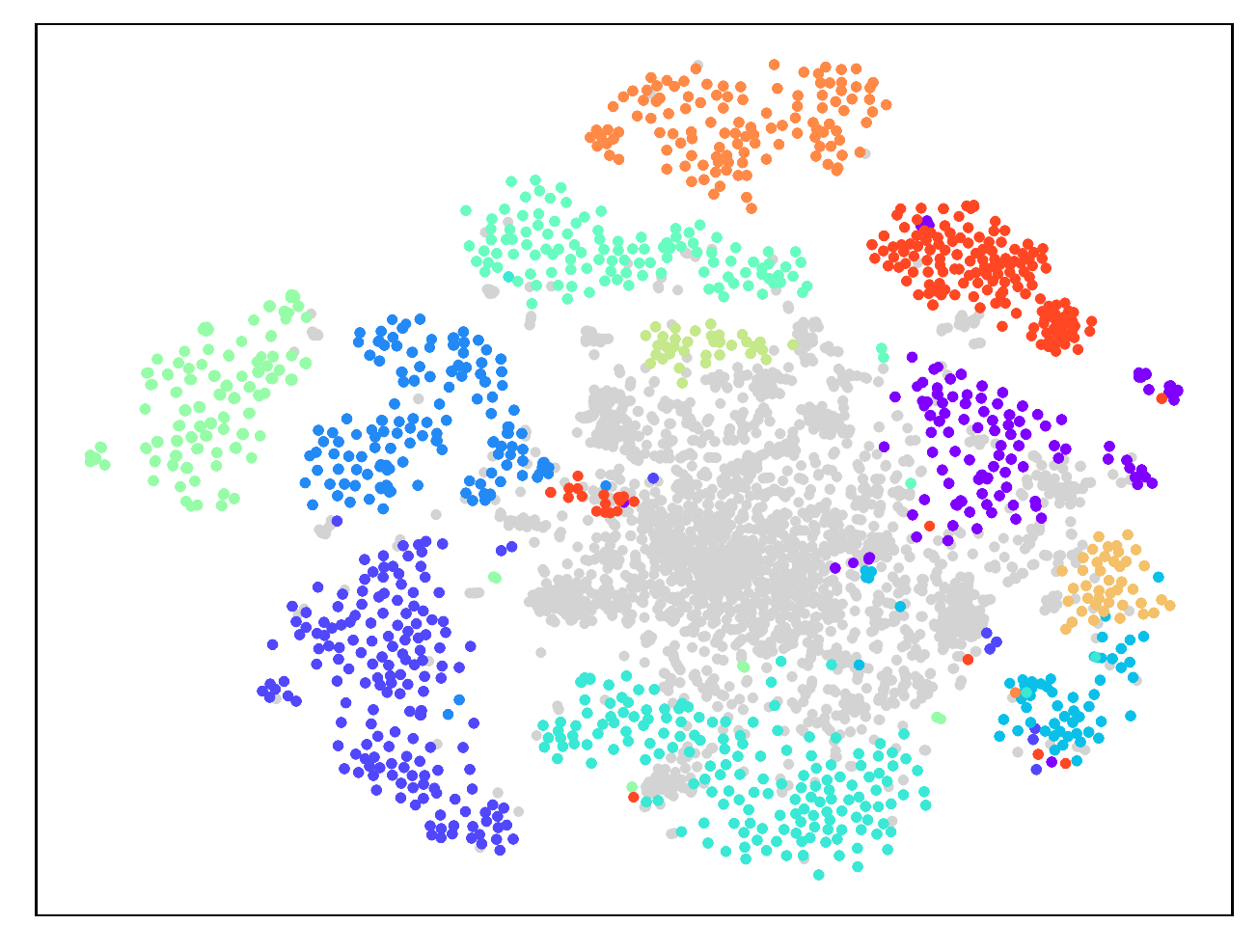}   &
	 \fig[.3]{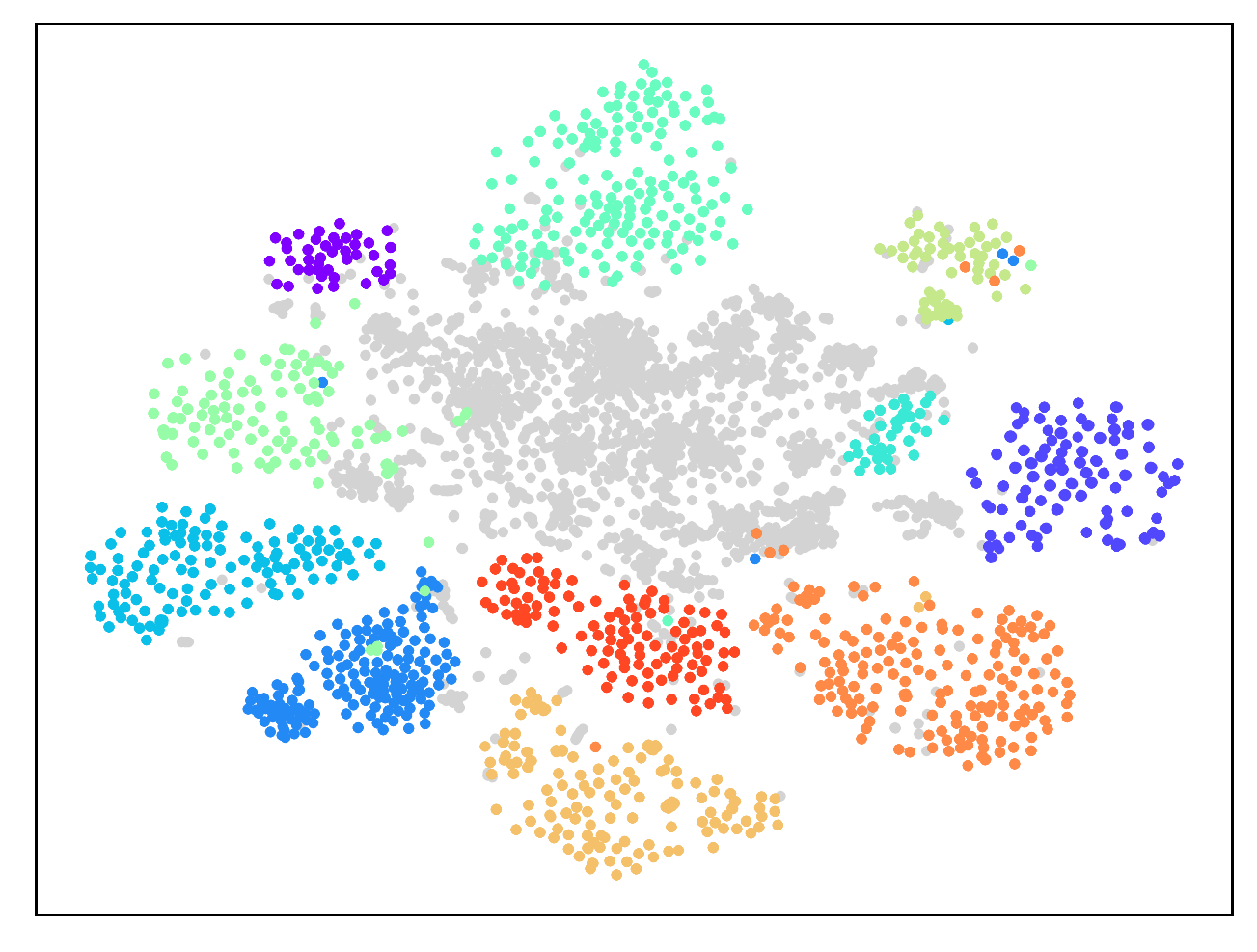}   &
	 \fig[.3]{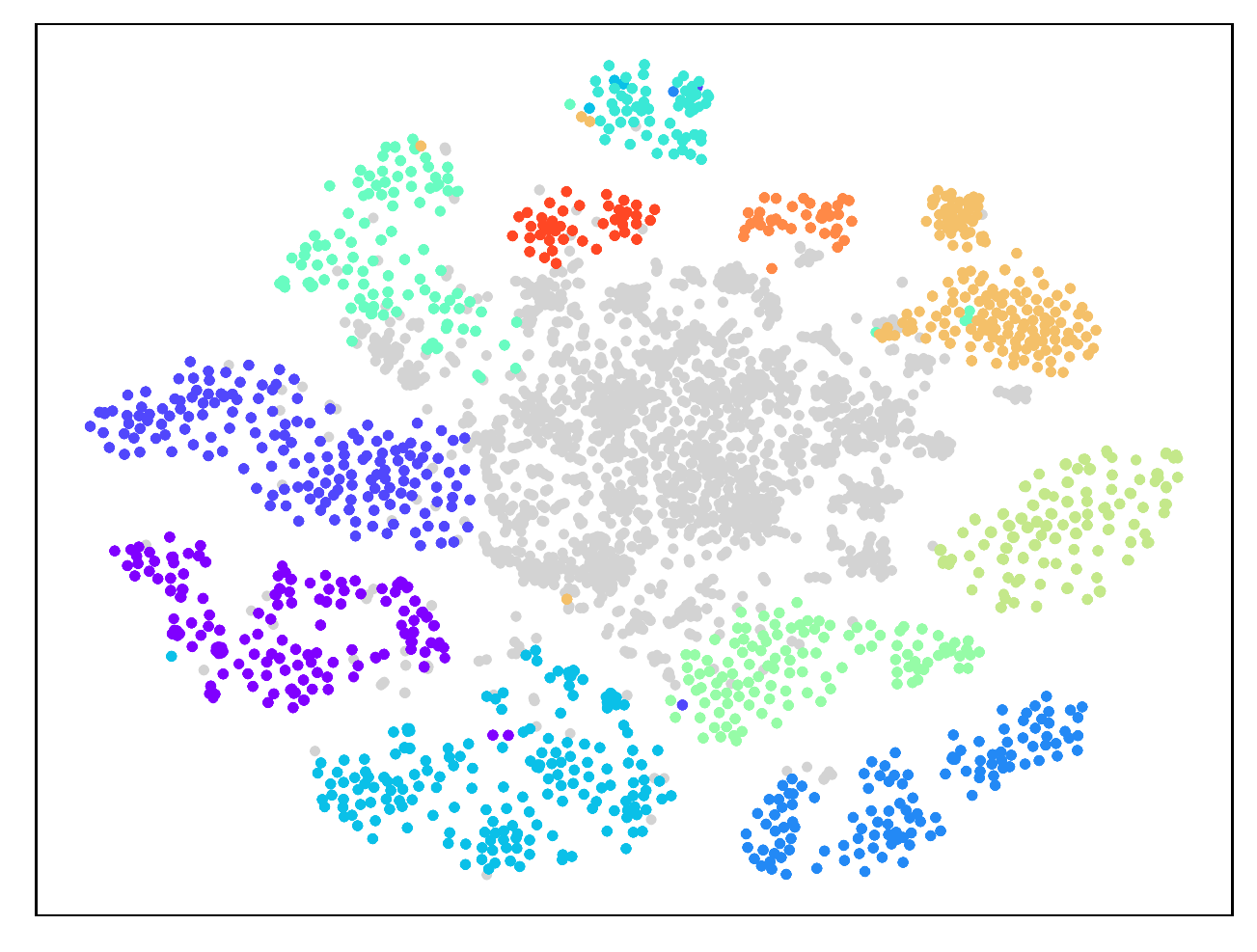}   \\[-2pt]
	 \raisebox{37pt}{Medium}                 &
	 \fig[.3]{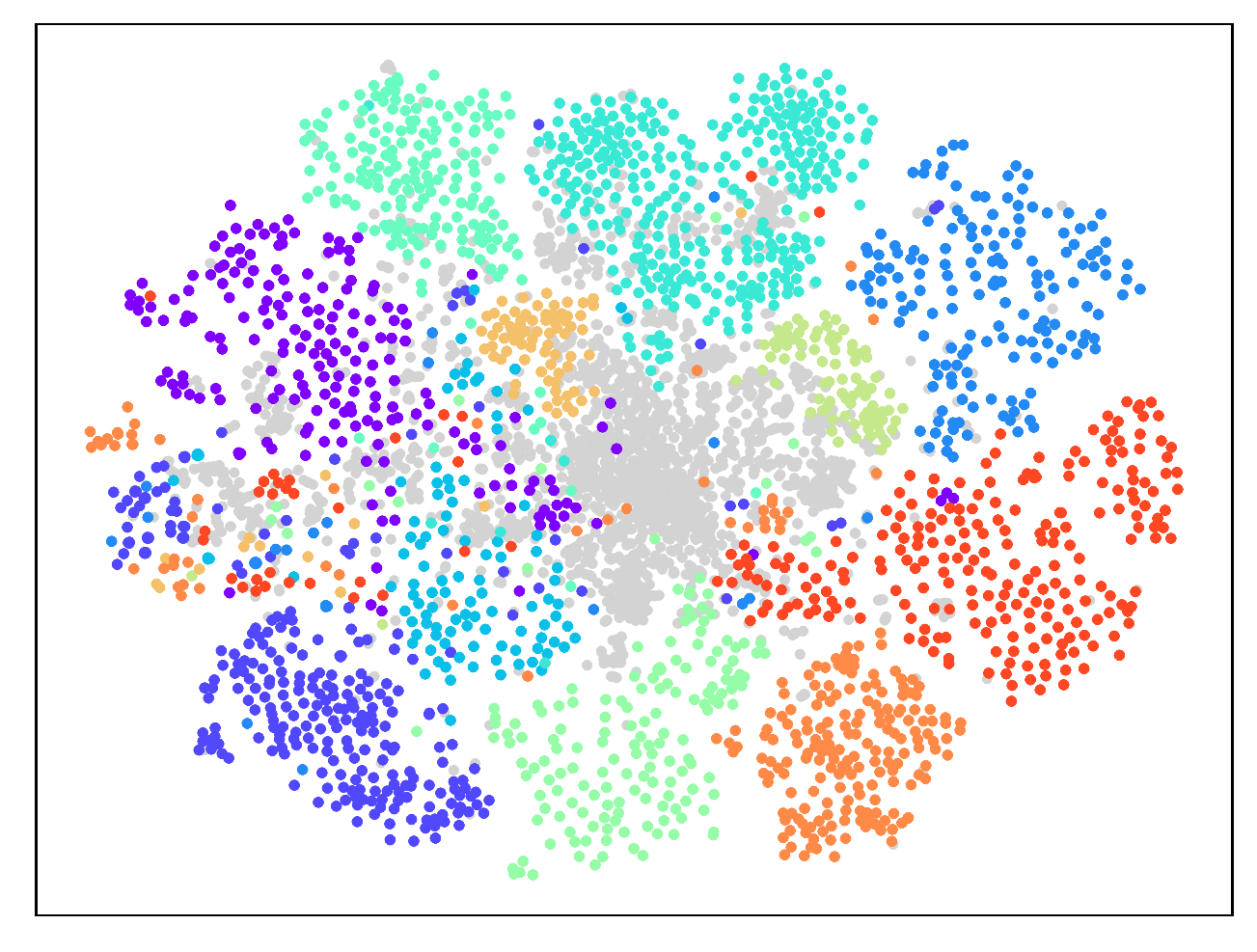} &
	 \fig[.3]{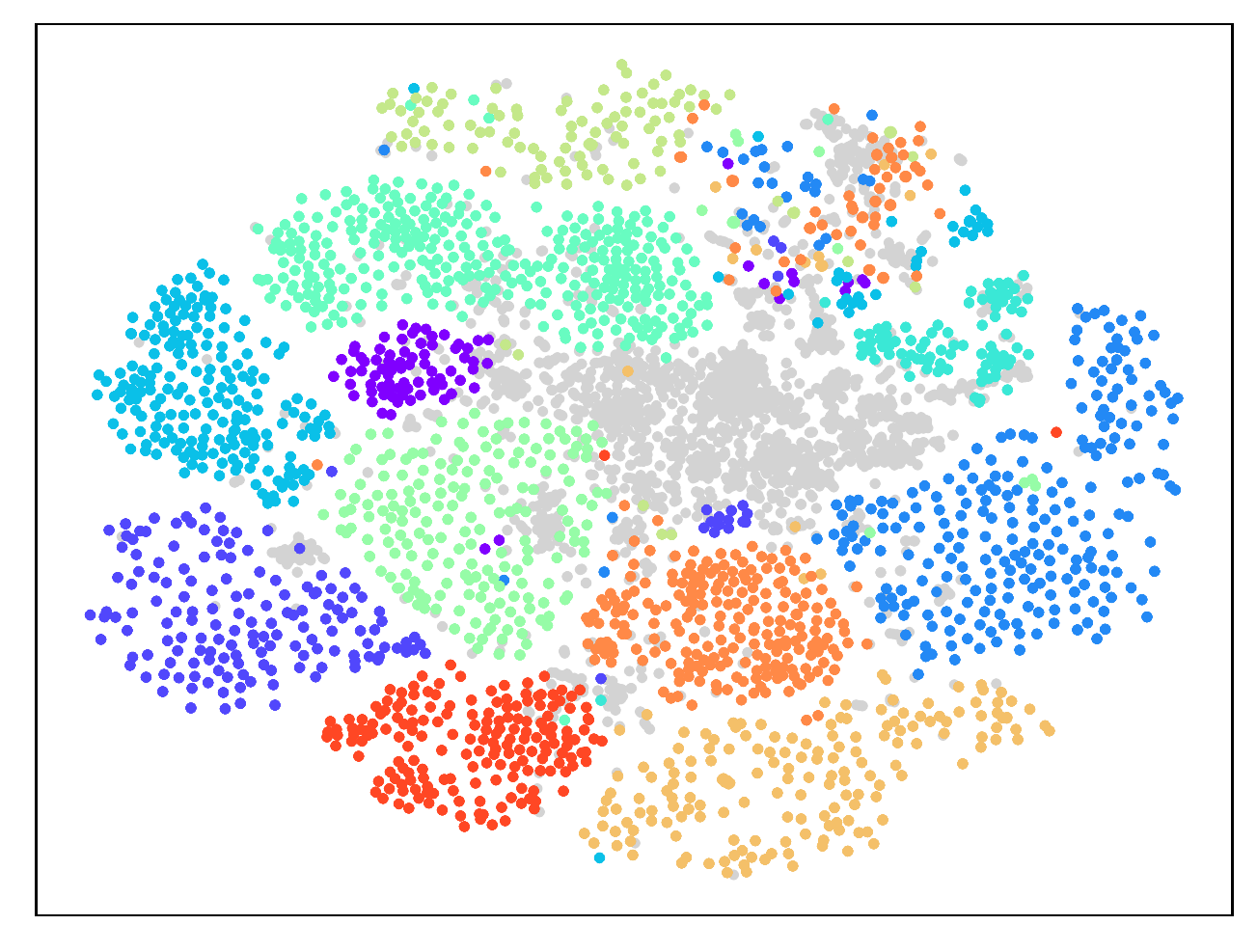} &
	 \fig[.3]{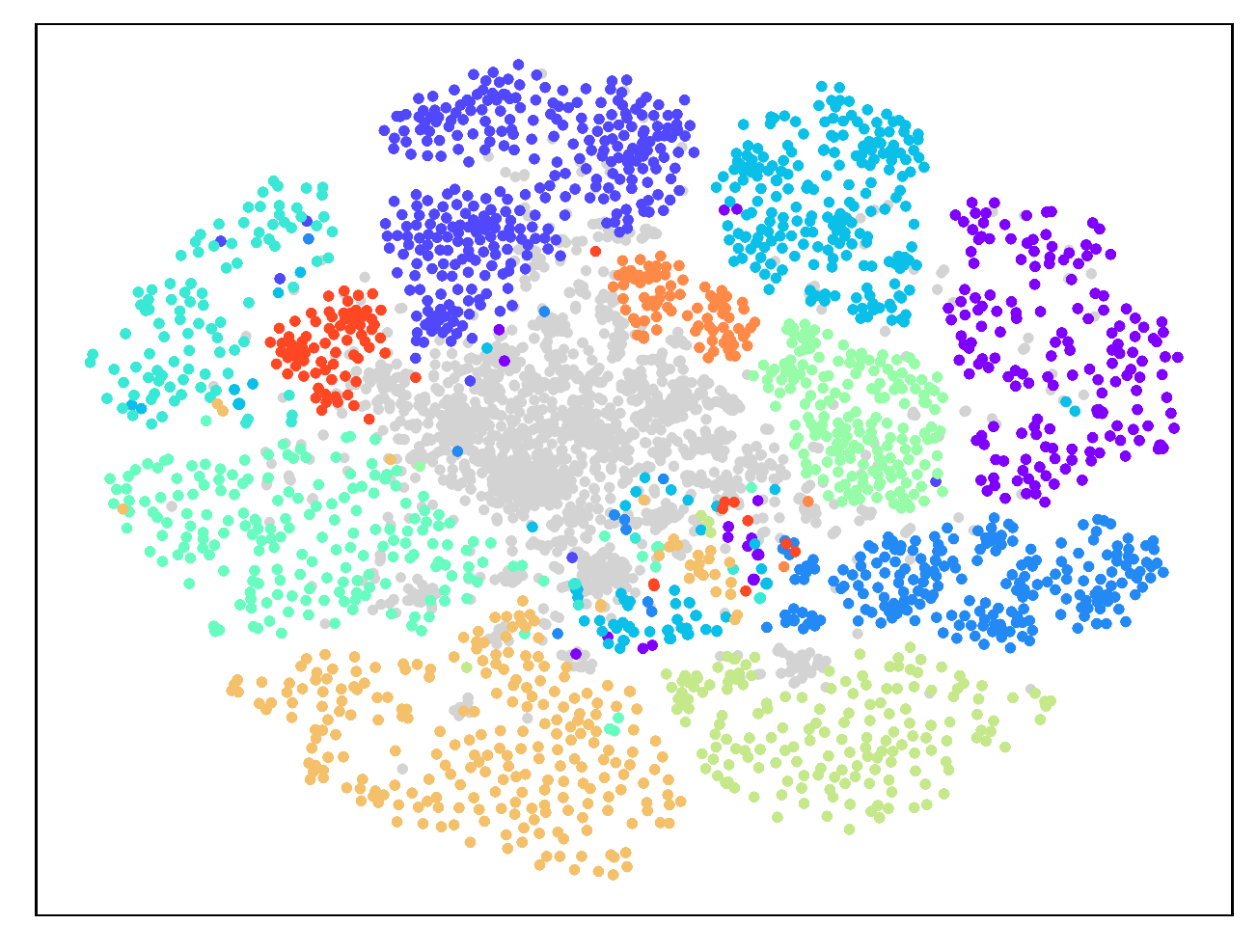} \\[-3pt]
	 \raisebox{37pt}{Hard}                   &
	 \fig[.3]{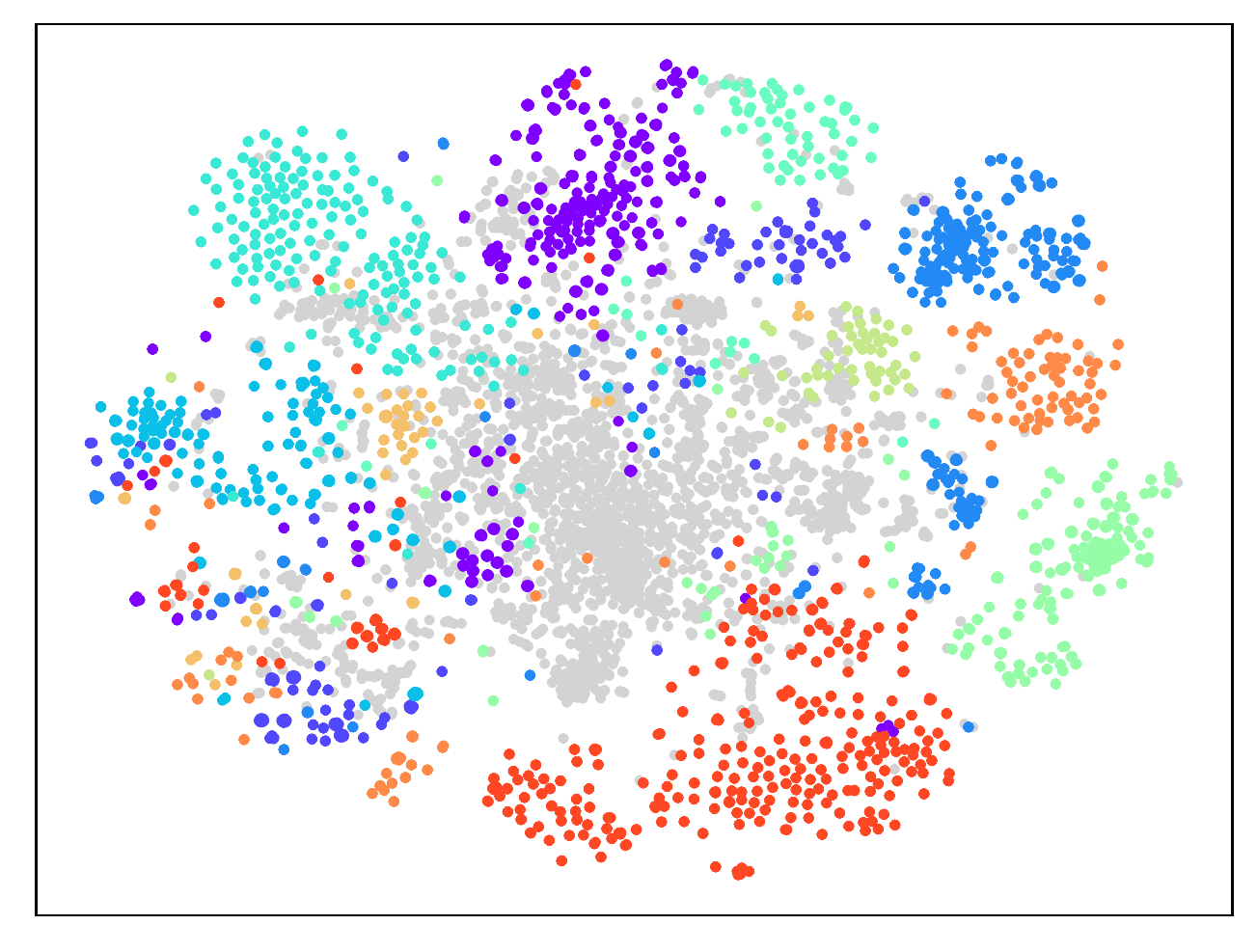}   &
	 \fig[.3]{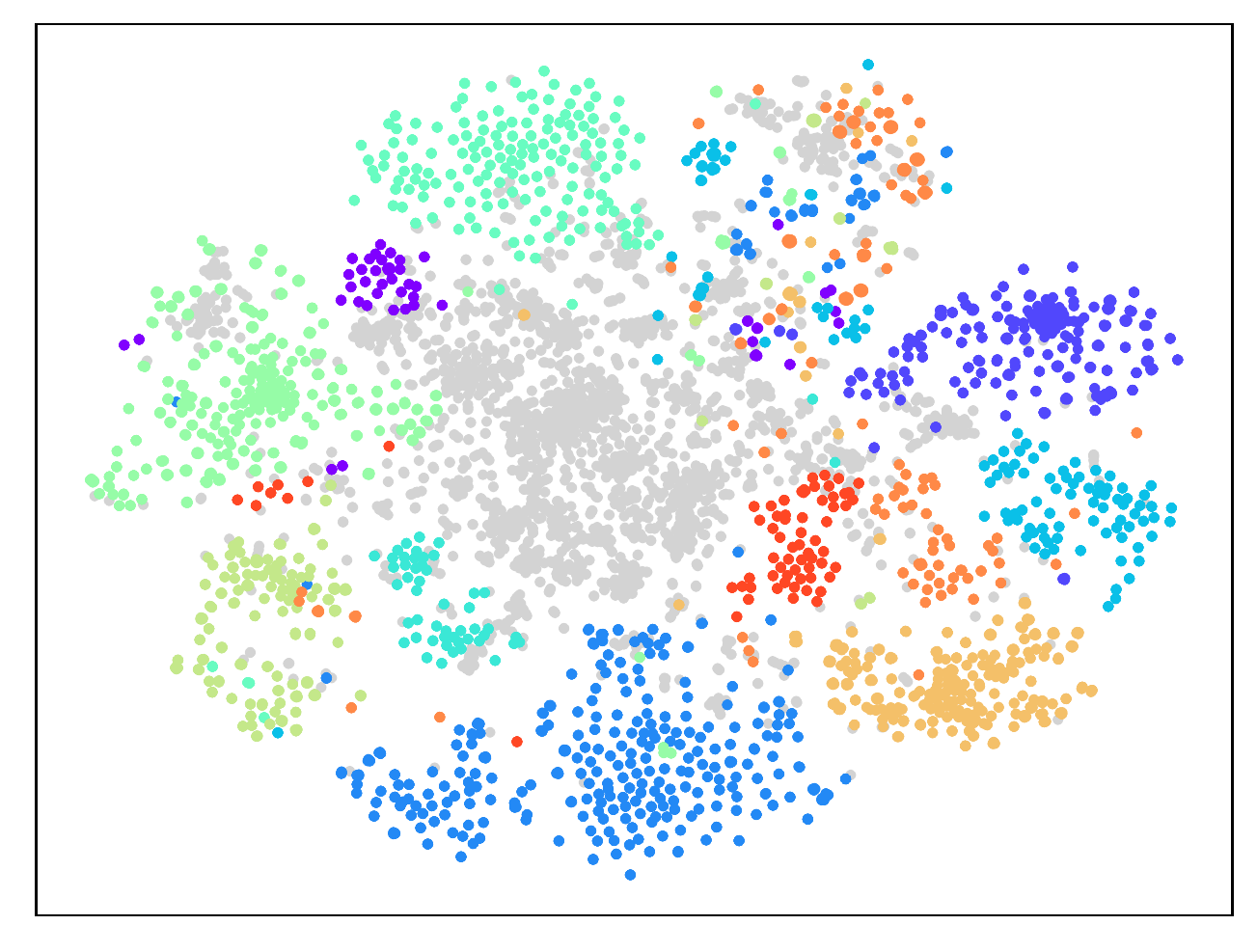}   &
	 \fig[.3]{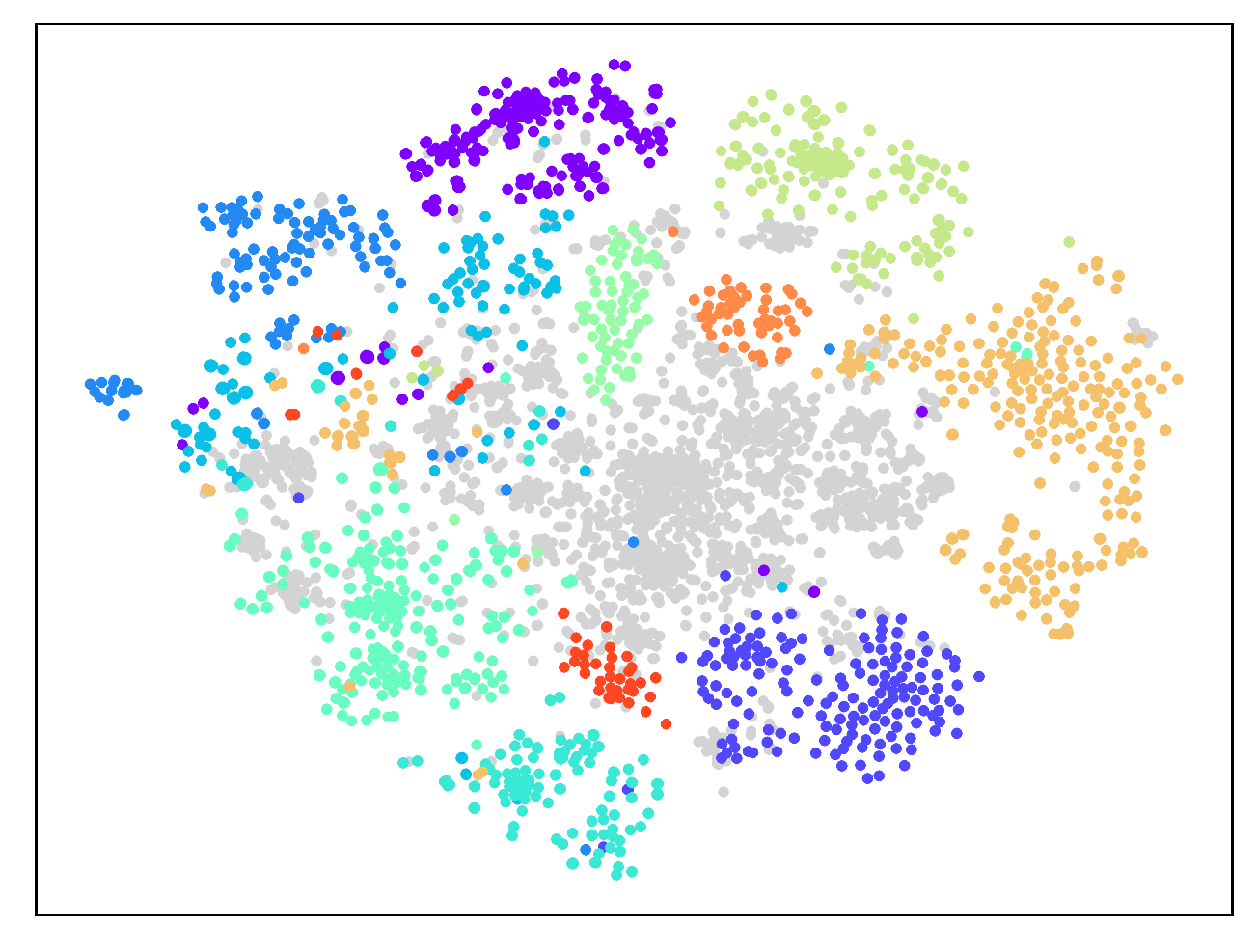}   \\
	                                         &
	 Resnet101                               &
	 R50+ViT-B baseline                      &
	 DToP-R50+ViT-B (ours)                   \\
\end{tabular}
\caption{Visualization of \emph{revisited Paris} (\rpar or \rpa) evaluation set under \emph{easy}, \emph{medium} and \emph{hard} protocols~\cite{RITAC18} (in rows) using
t-SNE on output embeddings~\eq{out} obtained by different models (in columns) fine-tuned on SfM-120k~\cite{Radenovic01}. For each protocol, positive images are colored by query group label and negative are gray. The set of medium positives is the union of easy and hard positives.}
\label{fig:tsne}
\end{figure}

\subsection{More visualizations}
\label{sec:more-vis}

\autoref{fig:tsne} provides
t-SNE
visualization of embeddings of \rpar~\cite{RITAC18} by different models trained on SfM-120k~\cite{Radenovic01}. It is clear that the class distribution of positive images under hard protocol are more overlapping than easy. Medium, being the union of easy and hard, is more populated but classes are better separated than in hard. Vision transformers clearly separate classes better than Resnet101, especially under hard protocol. The difference between our \ours-R50+ViT-B and the baseline hybrid model R50+ViT-B is small.